\documentclass[sigconf, nonacm]{acmart}

\AtBeginDocument{%
  }

\acmConference{}{}{}

\acmBooktitle{}

\usepackage{soul}
\usepackage{multirow}
\usepackage{subcaption}

\usepackage{float}
\usepackage{placeins}

\begin{document}


\title{Statistical vs. Deep Learning Models for Estimating Substance Overdose Excess Mortality in the US}


\author{Sukanya Krishna}
\email{sukanyakrishna@g.harvard.edu}
\affiliation{%
  \institution{Harvard University; Boston Children's Hospital}
  \city{Cambridge}
  \state{Massachusetts}
  \country{USA}
}

\author{Marie-Laure Charpignon}
\affiliation{%
  \institution{Boston Children's Hospital; Broad Institute of MIT and Harvard; Massachusetts Institute of Technology}
  \city{Cambridge}
  \state{Massachusetts}
  \country{USA}}

\author{Maimuna Majumder}
\affiliation{%
  \institution{Harvard Medical School; Boston Children's Hospital}
  \city{Cambridge}
  \state{Massachusetts}
  \country{USA}
}



\begin{abstract}
Substance overdose mortality in the United States claimed over 80,000 lives in 2023, with the COVID-19 pandemic exacerbating existing trends through healthcare disruptions and behavioral changes. Estimating excess mortality, defined as deaths beyond expected levels based on pre-pandemic patterns, is essential for understanding pandemic impacts and informing intervention strategies. However, traditional statistical methods like SARIMA assume linearity, stationarity, and fixed seasonality, which may not hold under structural disruptions.

We present a systematic comparison of SARIMA against three deep learning (DL) architectures (LSTM, Seq2Seq, and Transformer) for counterfactual mortality estimation using national CDC data (2015-2019 for training/validation, 2020-2023 for projection). We contribute empirical evidence that LSTM achieves superior point estimation (17.08\% MAPE vs. 23.88\% for SARIMA) and better-calibrated uncertainty (68.8\% vs. 47.9\% prediction interval coverage) when projecting under regime change. We also demonstrate that attention-based models (Seq2Seq, Transformer) underperform due to overfitting to historical means rather than capturing emergent trends. Our reproducible pipeline incorporates conformal prediction intervals and convergence analysis across 60+ trials per configuration, and we provide an open-source framework deployable with 15 state health departments. Our findings establish that carefully validated DL models can provide more reliable counterfactual estimates than traditional methods for public health planning, while highlighting the need for calibration techniques when deploying neural forecasting in high-stakes domains.
\end{abstract}






\maketitle


\section{Introduction}

The substance overdose crisis in the United States has reached epidemic proportions, claiming over 80,000 lives in 2023 alone \cite{cdc2023, us_overdose_2024_2025}. The trajectory of substance overdose mortality over the past two decades, characterized by successive waves driven by prescription opioids, heroin, synthetic fentanyl, and polysubstance use, exhibits temporal dependencies, spatial clustering, and community-level propagation patterns that parallel infectious disease (ID) outbreaks \cite{heuton_kapoor_shrestha_stopka_hughes_2024}.

The COVID-19 pandemic intensified this crisis through multiple mechanisms. Healthcare system disruptions reduced access to medication-assisted treatment and harm reduction services, social isolation increased substance use risks, and supply chain shifts toward more potent synthetic opioids drove fatality rates higher \cite{ghose_forati_mantsch_2022}. Understanding the magnitude of pandemic-attributable mortality requires estimating counterfactual trajectories, or projections of expected deaths had the pandemic not occurred. This estimation problem is distinct from forecasting: rather than predicting future values, we must construct plausible pre-pandemic trends that can be compared against observed outcomes to quantify excess mortality.

Traditional time series methods like Seasonal Autoregressive Integrated Moving Average (SARIMA) remain the standard approach in public health surveillance. However, SARIMA's core assumptions of linearity, stationarity over the forecast horizon, and time-invariant seasonal patterns may fail when applied to periods of structural change \cite{chandra2024}. For forecast horizons extending multiple years beyond training data, these assumptions become increasingly tenuous, particularly when the underlying data-generating process has shifted fundamentally. Deep learning (DL) models offer greater representational flexibility and have demonstrated superior performance in infectious disease forecasting tasks involving regime changes and nonlinear dynamics \cite{wu2020deeptransformermodelstime, venna_tavanaei_gottumukkala_raghavan_maida_nichols_2019}.

\subsection{Problem Formulation and Approach}

We formally define the counterfactual estimation problem as follows. Let $y_t$ represent observed monthly substance overdose deaths at time $t$, with training data $\{y_t\}_{t=1}^{T}$ spanning a stable pre-pandemic period. Our objective is to estimate $\hat{y}_{T+h}$ for projection horizon $h \in \{1, ..., H\}$, where $\hat{y}_{T+h}$ represents the counterfactual mortality level expected under continuation of pre-pandemic trends. Excess mortality is then computed as $\Delta_h = y_{T+h}^{\text{obs}} - \hat{y}_{T+h}$, where $y_{T+h}^{\text{obs}}$ are actual reported deaths during the pandemic period.

The core technical challenge lies in model selection and validation. Standard time series evaluation on held-out pandemic-period data would conflate model performance with the very structural breaks we aim to quantify. Instead, we validate models on the immediate pre-pandemic period (2019), selecting architectures that best extrapolate from earlier stable conditions (2015-2018). This validation strategy assumes that models demonstrating superior generalization on pre-pandemic data will produce more credible counterfactual projections.

We construct prediction intervals (PIs) using conformal prediction techniques adapted for time series, providing model-agnostic uncertainty quantification. A well-calibrated model should exhibit PI coverage near the nominal 95\% level on validation data, indicating appropriate uncertainty about out-of-sample predictions. During pandemic-period projection, we expect observed deaths to fall outside these intervals, and this deviation quantifies excess mortality while the interval width reflects projection uncertainty.

\subsection{Contributions}

Our work makes four primary contributions to the intersection of machine learning and computational health. First, we provide a systematic empirical comparison of traditional statistical methods (SARIMA) against modern DL architectures (LSTM, Seq2Seq, Transformer) for counterfactual mortality estimation. We offer the first rigorous benchmarking study comparing these approaches on national-level substance overdose data, with evaluation across multiple metrics balancing point estimation accuracy and uncertainty calibration. Second, we deliver methodological insights on architecture selection for counterfactual time series modeling. We demonstrate that LSTM models achieve superior performance (17.08\% MAPE vs. 23.88\% for SARIMA; 68.8\% vs. 47.9\% PI coverage) while attention-based architectures (Seq2Seq, Transformer) exhibit systematic overfitting to historical means. These findings challenge assumptions that newer, more complex architectures necessarily outperform recurrent models in low-data regimes. Third, we establish a rigorous uncertainty quantification framework incorporating conformal prediction intervals, multi-trial convergence analysis (60+ trials per configuration), and explicit calibration assessment. We show that DL models can achieve both improved point estimates and better-calibrated uncertainty bounds compared to traditional methods when properly validated and tuned. Fourth, we provide a reproducible open-source pipeline designed for deployment with state health departments. Our modular codebase, comprehensive documentation, and interactive dashboard lower barriers to adoption for public health practitioners, enabling local analysis with jurisdictional data while maintaining methodological rigor.

These contributions address a critical gap in substance use epidemiology, where most excess mortality studies rely exclusively on statistical models despite their known limitations under structural change. By systematically evaluating DL alternatives and providing actionable deployment tools, we enable more reliable counterfactual estimation to inform public health planning and resource allocation.

\section{Related Work}

Statistical models like SARIMA remain widely used in public health surveillance \cite{held_höhle_hofmann_2005, shaman_karspeck_2012}, valued for interpretability and computational efficiency. However, recent work has questioned their suitability for capturing complex epidemic dynamics. Chandra et al. (2024) used SARIMA to estimate over 25,000 excess substance overdose deaths during COVID-19, but reported prediction intervals ranging from 2,800 to 48,500, a span that undermines practical utility \cite{chandra2024, widecis}. This uncertainty reflects SARIMA's difficulty handling the nonlinear interactions and regime shifts characteristic of substance use patterns.

Given structural similarities between substance overdose trends and infectious disease outbreaks (including temporal dependencies, spatial clustering, and behavioral feedback loops), methods from epidemic forecasting offer relevant precedents. LSTM and Seq2Seq models have consistently outperformed statistical baselines for influenza prediction, particularly at longer horizons \cite{venna_tavanaei_gottumukkala_raghavan_maida_nichols_2019, Zhu_Fu_Yang_Ma_Hao_Chen_Liu_Li_Liu_Guo_et_2019}. Wu et al. (2020) demonstrated that Transformer architectures with self-attention mechanisms can capture global temporal dependencies in influenza-like illness (ILI) data \cite{wu2020deeptransformermodelstime}, though subsequent work has questioned whether attention provides consistent advantages over simpler recurrent architectures in univariate settings.

Compartmental epidemic models (e.g., SIR-based frameworks) have been adapted for substance use, incorporating treatment entry/exit, relapse dynamics, and social network effects \cite{Bottcher2024, blanco_wiley_lloyd_lopez_volkow_2020, battista_pearcy_strickland_2019}. While valuable for mechanistic understanding, these models require detailed parameterization of transition rates and often lack the flexibility to capture unanticipated regime changes. Recent work has explored hybrid approaches combining mechanistic structure with learned components, though application to substance overdose remains limited \cite{bansback_barbosa_barocas_bayoumi_behrends_chhatwal_cipriano_coffin_goldhaber-fiebert_hoch_2021}.

Foundation models and large-scale pretraining represent an emerging direction. TimeGPT and similar approaches learn temporal representations from diverse time series, potentially offering improved generalization through transfer learning \cite{comparativedeeplearningsubover}. However, these models typically require substantial GPU memory for fine-tuning and may not provide transparent uncertainty quantification, both critical requirements for public health deployment. Our work focuses on architectures trainable from scratch with limited data, prioritizing accessibility for resource-constrained health departments while maintaining rigorous uncertainty bounds.

Despite growing interest in DL for epidemic forecasting, few studies have systematically compared architectures for counterfactual estimation in substance use epidemiology. Most overdose modeling efforts remain focused on geographic simulation (agent-based models, spatial diffusion) or rely exclusively on traditional statistics \cite{comparativedeeplearningsubover}. We address this gap through comprehensive benchmarking on national mortality data with emphasis on both accuracy and calibrated uncertainty.

\section{Methodology}

\begin{table*}[htbp]
\centering
\caption{Optimal hyperparameter settings selected based on validation RMSE.}
\begin{tabular}{@{}lllll@{}}
\toprule
\textbf{Model} & \textbf{Lookback} & \textbf{Batch Size} & \textbf{Epochs} & \textbf{Notes} \\
\midrule
SARIMA         & ---    & ---    & ---  & (1,0,0)(1,1,1,12) selected via grid search \\
LSTM           & 5     & 8     & 50  & 2-layer LSTM with ReLU activation \\
Seq2Seq   & 7     & 16     & 100   & GRU 64 encoder - 64 decoder without attention \\
Seq2Seq w/ Attn.   & 5     & 16     & 50   & GRU 128 encoder - 64 decoder with Bahdanau attention \\
Transformer    & 7     & 32     & 100  & $d=64$, 2-head self-attention, positional encodings \\
\bottomrule
\end{tabular}
\label{hyperparameter}
\end{table*}

\subsection{Data Sources and Extraction}

We use final substance overdose mortality data from the Centers for Disease Control and Prevention (CDC) Wide-ranging Online Data for Epidemiologic Research (CDC WONDER) Multiple Cause of Death database. Monthly counts of national-level deaths spanning January 2015-December 2023 are extracted using ICD-10 codes for underlying causes (X40-X45: accidental poisoning; X60-X65: intentional self-poisoning; X85: assault; Y10-Y15: undetermined intent) combined with contributing cause codes for specific substances (T40.1: heroin; T40.2: natural/semisynthetic opioids; T40.4: synthetic opioids excluding methadone; T40.5: cocaine; T43.6: psychostimulants; T42.4: benzodiazepines; T51: alcohol). 

This extraction protocol captures deaths involving opioids, stimulants, benzodiazepines, and alcohol, which are the primary substance classes driving US overdose mortality. We acknowledge that "substance use" encompasses broader phenomena, but restrict our analysis to these documented causes as recorded in CDC WONDER. Detailed extraction procedures and ICD-10 code mappings are provided in Appendix A.

We define "pandemic-period" as calendar years 2020-2023, recognizing that while the official US public health emergency spanned March 2020-May 2023, substance use patterns and healthcare access were disrupted throughout this broader window.

\subsection{Dataset Split and Validation Strategy}

We partition data into three temporally contiguous periods: training from January 2015 through December 2018 (48 months), validation from January through December 2019 (12 months), and projection from January 2020 through December 2023 (48 months). This structure enables assessment of generalization from stable training conditions (2015-2018) to a stable but held-out pre-pandemic year (2019). The validation period serves as our primary model selection criterion: we hypothesize that architectures performing well on 2019 data, when no structural breaks had yet occurred, will produce more credible counterfactual projections for 2020-2023.

Critically, no model receives information from 2020-2023 during training or hyperparameter tuning. The projection period is used exclusively for post-hoc evaluation of counterfactual plausibility, not for accuracy measurement in the traditional forecasting sense. This design acknowledges that "accuracy" is undefined for counterfactual scenarios; instead, we assess whether projections follow plausible pre-pandemic trend extrapolations and exhibit appropriate uncertainty calibration.

\begin{table*}[htbp]
\centering
\caption{Model Performance over Training Period (2015-2018) and Validation Period (2019).}
\begin{tabular}{@{}lcccccccc@{}}
\toprule
\multirow{2}{*}{\textbf{Model}} &
\multicolumn{4}{c}{\textbf{Training (2015–2018)}} &
\multicolumn{4}{c}{\textbf{Validation (2019)}} \\
\cmidrule(lr){2-5} \cmidrule(lr){6-9}
& RMSE & MAE & MAPE & PI Cov. &
  RMSE & MAE & MAPE & PI Cov. \\
\midrule
SARIMA & 779.50 ± 0.00 & 343.75 ± 0.00 & 6.69\% ± 0.00 & 100\% &
        402.25 ± 0.00 & 304.24 ± 0.00 & 4.84\% ± 0.00 & 53.65\% \\
LSTM   & 314.58 ± 9.32 & 241.47 ± 5.40 & 4.37\% ± 0.10 & 100.0\% &
        \textbf{293.13 ± 27.61} & \textbf{248.33 ± 26.22} & \textbf{4.04\% ± 0.41} & \textbf{96.67\%} \\
Seq2Seq & 244.76 ± 0.59 & 189.34 ± 0.26 & 3.28\% ± 0.01 & 100.0\% &
         216.70 ± 1.58 & 182.90 ± 1.60 & 3.04\% ± 0.02 & 76.6\% \\
Seq2Seq w/ Attn. & 244.54 ± 0.99 & 186.41 ± 0.83 & 3.23\% ± 0.02 & 96.8\% &
                  233.67 ± 24.57 & 197.35 ± 20.86 & 3.26\% ± 0.31 & 89.13\% \\
Transformer & 254.04 ± 26.52 & 202.35 ± 18.75 & 3.58\% ± 0.36 & 98.15\% &
             334.18 ± 75.22 & 285.01 ± 69.33 & 4.63\% ± 1.09 & 90.17\% \\
\bottomrule
\end{tabular}
\label{tab:prepandemic_tuning_metrics}
\end{table*}

\subsection{Models}

We evaluate four model classes spanning traditional statistics and modern deep learning:

\subsubsection{SARIMA (Baseline)}
Seasonal ARIMA serves as our statistical baseline, widely adopted in public health for its interpretability and established theory \cite{predictive_models, sarima}. We specify SARIMA$(p,d,q)(P,D,Q,s)$ where $(p,d,q)$ are non-seasonal autoregressive order, differencing degree, and moving average order; $(P,D,Q)$ are seasonal equivalents; and $s=12$ reflects monthly periodicity. 

We conduct exhaustive grid search over $p, d, q, P, D, Q \in \{0,1,2\}$, training 30 instances per configuration with different random initializations. The configuration minimizing validation RMSE is selected. This approach mirrors established SARIMA tuning protocols from influenza and COVID-19 surveillance studies \cite{predictive_models}.

\subsubsection{LSTM}
Long Short-Term Memory networks address SARIMA's linearity limitations through gated recurrent units capable of learning long-range dependencies and nonlinear relationships \cite{lstm, wu2020deeptransformermodelstime}. Our architecture uses two stacked LSTM layers (hidden dimension 64-128, tuned via grid search) followed by a fully connected output layer with ReLU activation. This design balances expressiveness with trainability on limited data (48 training months).

Input sequences are constructed via sliding windows: for lookback length $L$, we create batches $\{(y_{t-L:t}, y_{t+1})\}$ where $y_{t-L:t}$ is the input sequence and $y_{t+1}$ the target. Autoregressive inference generates multi-step projections by feeding predictions back as inputs.

\subsubsection{Seq2Seq}
We implement GRU-based encoder-decoder architectures with optional Bahdanau attention \cite{bahdanau2016neuralmachinetranslationjointly, seq2seq}. The encoder processes input sequences into fixed-length context vectors; the decoder autoregressively generates output sequences. Attention mechanisms allow decoders to selectively focus on relevant encoder states, potentially improving long-horizon projection quality.

We evaluate both attention-enabled and attention-free variants, with encoder/decoder hidden dimensions {64, 128} tuned separately. This design tests whether attention provides meaningful advantages for univariate mortality series, or if simpler recurrent encoding suffices.

\subsubsection{Transformer}
Inspired by univariate time series forecasting adaptations \cite{wu2020deeptransformermodelstime, covtransformer}, our Transformer uses single-head or multi-head self-attention (1-2 heads, tuned) with sinusoidal positional encodings and a feedforward output layer. Unlike multivariate Transformers (e.g., Temporal Fusion Transformer), our architecture receives only historical death counts as input, without any external covariates.

This streamlined design tests whether self-attention's ability to capture global temporal dependencies translates to improved counterfactual estimation in data-limited settings. We use autoregressive decoding for consistency with Seq2Seq evaluation.

\subsection{Hyperparameter Tuning}

All deep learning models undergo structured grid search over lookback windows of 3, 5, 7, 9, 11, and 12 months, batch sizes of 8, 16, and 32, and training epochs of 50 and 100. For Seq2Seq models, we additionally tune encoder and decoder hidden dimensions across values of 64 and 128, and evaluate both with and without attention mechanisms. For Transformer models, we test 1 or 2 attention heads with a fixed hidden dimension of 64.

For each configuration, we train 30 trials with fixed random seed (42 for main results; sensitivity analysis across seeds in Appendix B). Validation RMSE averaged over trials determines optimal hyperparameters. This multi-trial approach accounts for optimization stochasticity inherent to neural network training.

Selected configurations are shown in Table~\ref{hyperparameter}. Models are implemented in PyTorch and trained on GPU-enabled infrastructure. Detailed hyperparameter search logs and convergence plots are available in our public repository.

\subsection{Uncertainty Quantification}

Beyond point estimates, we compute 95\% prediction intervals (PIs) using model-specific strategies. For SARIMA, we use analytical PIs derived from residual variance estimates and forecast error propagation, as implemented in standard statistical libraries. For deep learning models, we construct conformal prediction intervals via quantile regression on validation residuals. For each validation prediction $\hat{y}_t$, we compute residual $r_t = |y_t - \hat{y}_t|$. The 95th percentile of validation residuals, $q_{0.95}$, defines the PI for projection timestep $h$ as $[\hat{y}_{T+h} - q_{0.95}, \hat{y}_{T+h} + q_{0.95}]$. This approach provides distribution-free uncertainty bounds without assuming Gaussian residuals.

We evaluate PI quality through two metrics: coverage, measured as the proportion of validation or projection observations falling within computed intervals (well-calibrated models should achieve approximately 95\% coverage on validation data), and width, measured as the average interval span and indicating uncertainty magnitude. Narrower intervals indicate greater confidence but risk undercoverage if overconfident. Prediction intervals differ from confidence intervals in that PIs account for both parameter uncertainty and inherent outcome variability, representing the range where individual future observations are expected to fall. This distinction is critical when projecting mortality counts, where month-to-month variation is substantial.

\subsection{Evaluation Metrics}

We assess models using six complementary metrics capturing both point estimation accuracy and uncertainty calibration. For point accuracy, we compute Root Mean Squared Error (RMSE), Mean Absolute Error (MAE), and Mean Absolute Percentage Error (MAPE) using standard formulations. For uncertainty quality, we evaluate prediction interval coverage (the proportion of observations within 95\% PIs), PI width (average span of prediction intervals), and overlap with observations (degree to which pandemic-period observations fall outside projected intervals, quantifying detected excess mortality).

On validation data (2019), accuracy metrics measure generalization quality under stable conditions. On projection data (2020-2023), these metrics assess counterfactual plausibility rather than traditional accuracy, since observed outcomes reflect structural disruptions we aim to quantify through excess mortality calculations.

\subsection{Experimental Protocol}

Our experimental workflow proceeds in three stages. During hyperparameter selection, we train each model class on 2015-2018 data across all hyperparameter combinations, evaluate on the 2019 validation set using RMSE, MAE, MAPE, and PI coverage, and select the configuration minimizing validation RMSE for each architecture. We repeat this process 30 times per configuration to account for initialization variance. For final model training, we retrain optimal configurations on combined 2015-2019 data, generate counterfactual projections for 2020-2023, compute prediction intervals using 2019 residuals, and assess projection plausibility and calibration. Finally, for excess mortality estimation, we calculate monthly excess deaths as $\Delta_h = y_{T+h}^{\text{obs}} - \hat{y}_{T+h}$, aggregate excess mortality over the pandemic period, and quantify sensitivity to model choice via cross-architecture comparison.

This protocol ensures that model selection relies exclusively on pre-pandemic performance, projections represent genuine extrapolation rather than interpolation, and uncertainty estimates reflect realistic out-of-sample variability. By validating on stable 2019 data, we identify models best able to capture underlying mortality dynamics when no regime change has occurred, providing our best proxy for credible counterfactual projection capability.

\section{Experiments and Analysis}

\subsection{Validation Performance}

Table~\ref{tab:prepandemic_tuning_metrics} reports performance on the 2015-2018 training period and 2019 validation period for all models. These results guide model selection for counterfactual projection.

On the critical 2019 validation period, LSTM achieves the best overall performance: lowest RMSE (293.13), MAE (248.33), MAPE (4.04\%), and highest PI coverage (96.67\%). This demonstrates superior generalization from 2015-2018 training data to held-out 2019 conditions, exactly the capability required for credible counterfactual projection.

Both Seq2Seq variants achieve competitive point estimates but with substantially degraded PI coverage (76.6\%-89.13\%), indicating overconfidence in predictions. The Transformer model shows the weakest validation performance (334.18 RMSE, 90.17\% coverage), despite reasonable training fit. This suggests overfitting to training period patterns that don't generalize forward.

SARIMA demonstrates the poorest validation accuracy (402.25 RMSE, 4.84\% MAPE) and severely undercovers (53.65\%), with prediction intervals failing to capture more than half of 2019 observations despite nominal 95\% confidence. This miscalibration raises concerns about SARIMA's reliability for pandemic-period projection, where structural change makes extrapolation even more challenging.

Based on these validation results, we select LSTM as our primary architecture for counterfactual estimation, given its superior accuracy-calibration balance. We proceed with all models to assess sensitivity of excess mortality estimates to model choice.

\begin{table*}[htbp]
\centering
\caption{Model Performance over Full Training Period (2015-2019) and Projection Period (2020-2023)}
\resizebox{\textwidth}{!}{
\begin{tabular}{@{}lcccccccc@{}}
\toprule
\multirow{2}{*}{\textbf{Model}} &
\multicolumn{4}{c}{\textbf{Training (2015–2019)}} &
\multicolumn{4}{c}{\textbf{Projection (2020–2023)}} \\
\cmidrule(lr){2-5} \cmidrule(lr){6-9}
& RMSE & MAE & MAPE & PI Cov. &
  RMSE & MAE & MAPE & PI Cov. \\
\midrule
SARIMA         & 553.69 ± 468.51 & 295.08 ± 468.51 & 5.58\% ± 9.56 & 98.2\% &
                 2.26e3 ± 716.09 & 2.15e3 ± 716.09 & 23.88\% ± 6.67 & 47.9\% \\
LSTM           & 324.74 ± 215.03 & 243.36 ± 215.03 & 4.29\% ± 3.74 & 100.0\% &
                 \textbf{1.70e3 ± 714.47} & \textbf{1.54e3 ± 714.47} & \textbf{17.08\% ± 7.21} & \textbf{68.8\%} \\
Seq2Seq        & 269.45 ± 161.26 & 215.87 ± 161.26 & 3.82\% ± 2.91 & 100.0\% &
                 2.64e3 ± 821.75 & 2.51e3 ± 821.75 & 27.88\% ± 7.61 & 45.8\% \\
Seq2Seq w/ Attn. & 269.27 ± 165.48 & 212.42 ± 165.48 & 3.78\% ± 2.99 & 100.0\% &
                   2.54e3 ± 797.76 & 2.41e3 ± 797.76 & 26.83\% ± 7.38 & 47.9\% \\
Transformer    & 322.18 ± 197.43 & 254.60 ± 197.43 & 4.53\% ± 3.71 & 100.0\% &
                 2.98e3 ± 832.01 & 2.86e3 ± 832.01 & 31.97\% ± 7.35 & 43.8\% \\
\bottomrule
\end{tabular}}
\label{tab:train_projection_compact}
\end{table*}

\begin{figure*}[ht]
  \centering
  \begin{subfigure}[b]{0.48\textwidth}
    \centering
    \includegraphics[width=\textwidth]{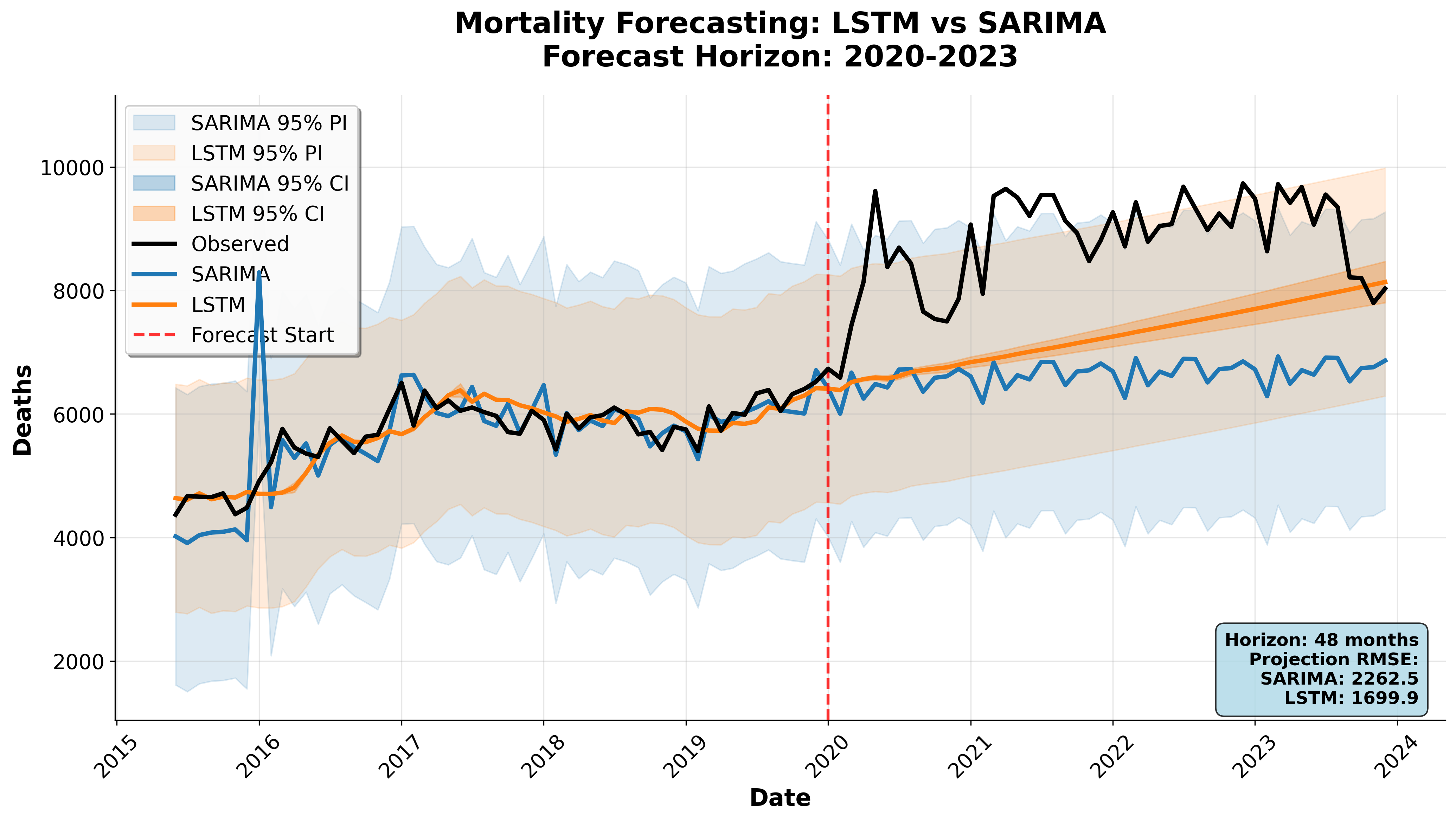}
    \caption{SARIMA vs LSTM}
    \label{fig:sarima_vs_lstm}
  \end{subfigure}%
  \hfill
  \begin{subfigure}[b]{0.48\textwidth}
    \centering
    \includegraphics[width=\textwidth]{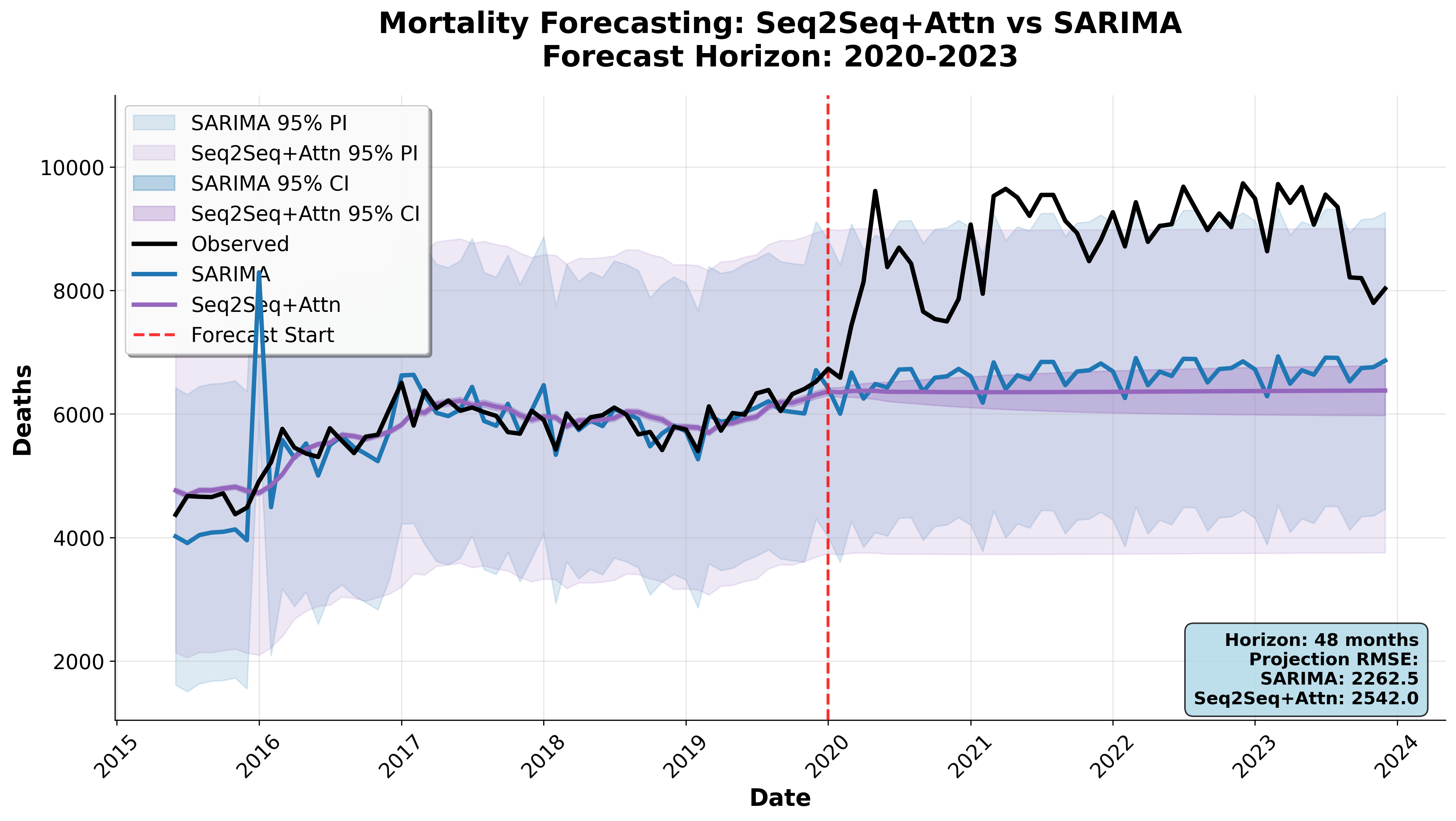}
    \caption{SARIMA vs Seq2Seq (w/ Attention)}
    \label{fig:sarima_vs_seq2seq}
  \end{subfigure}
  
  \vspace{0.5\baselineskip}
  
  \begin{subfigure}[b]{0.48\textwidth}
    \centering
    \includegraphics[width=\textwidth]{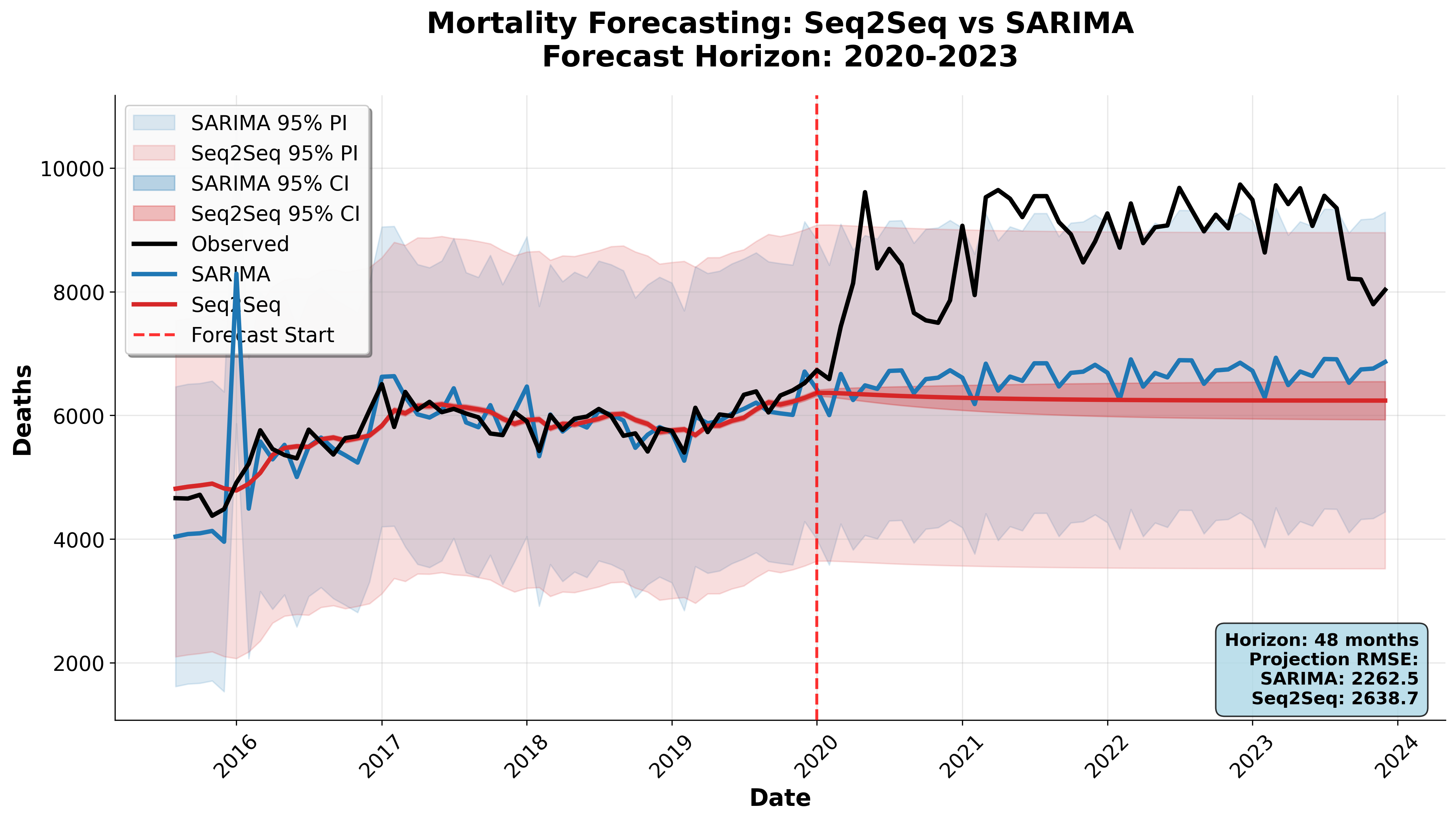}
    \caption{SARIMA vs Seq2Seq (w/out Attention)}
    \label{fig:sarima_vs_tcn}
  \end{subfigure}%
  \hfill
  \begin{subfigure}[b]{0.48\textwidth}
    \centering
    \includegraphics[width=\textwidth]{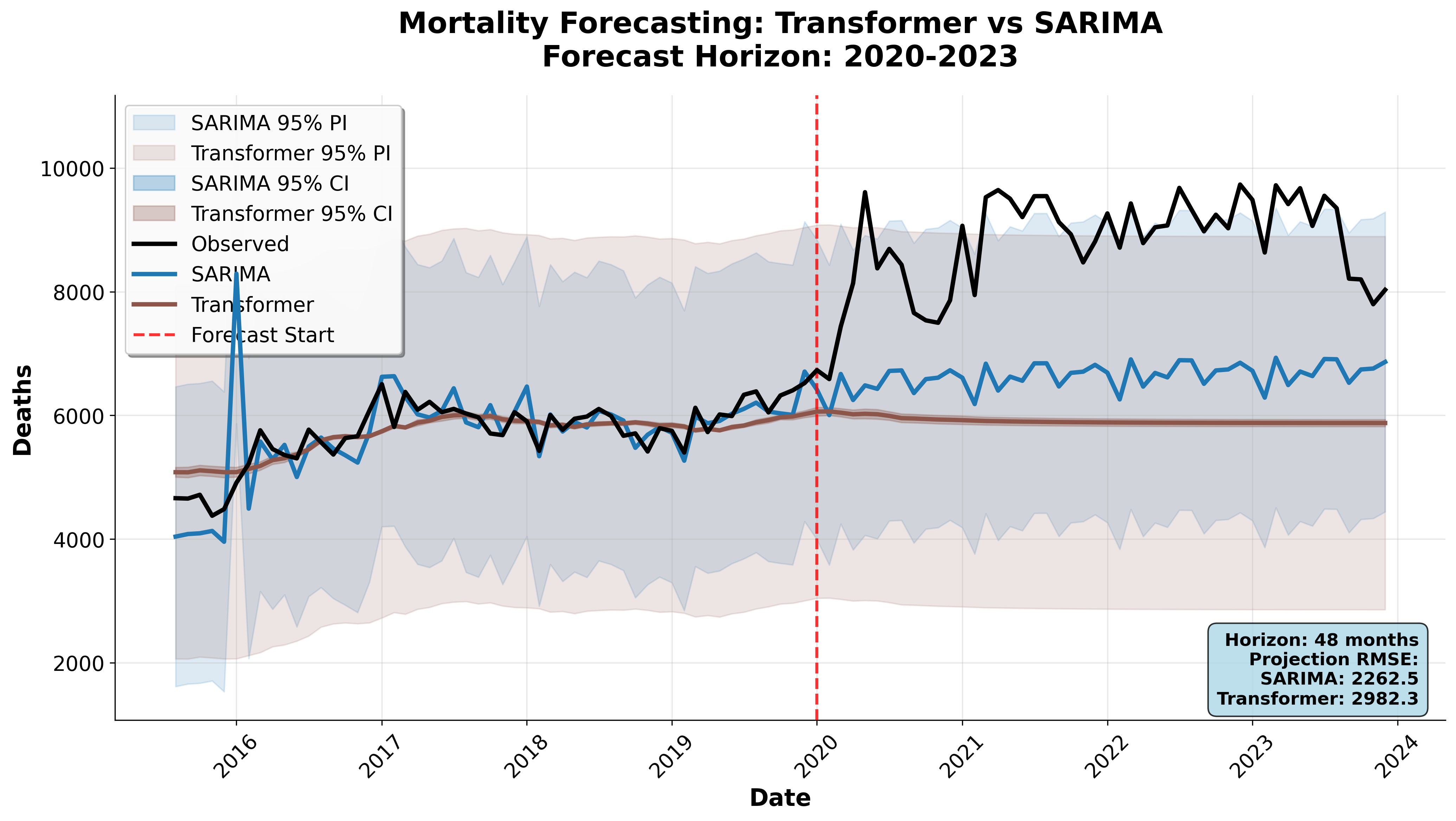}
    \caption{SARIMA vs Transformer}
    \label{fig:sarima_vs_transformer}
  \end{subfigure}
  
  \caption{Counterfactual mortality projections comparing SARIMA (baseline) against four DL architectures. Black line shows reported monthly substance overdose deaths; colored lines show model projections; shaded regions indicate 95\% prediction/confidence intervals. Vertical dashed line (January 2020) marks projection start. LSTM (a) tracks the rising trend visible in late 2019, while attention-based models (b-d) project flat trajectories that underestimate observed mortality.}
  \label{fig:forecast_comparisons}
\end{figure*}

\subsection{Counterfactual Projection Analysis}

Table~\ref{tab:train_projection_compact} reports performance when models trained on 2015-2019 data are projected forward through 2020-2023. We emphasize that these "errors" measure deviation from observed pandemic-period outcomes, not traditional forecast accuracy. The goal is to quantify excess mortality, not predict disrupted trends.

LSTM maintains the best projection metrics: lowest deviation from observed deaths (1.70e3 RMSE, 17.08\% MAPE) and highest PI coverage (68.8\%). The 68.8\% coverage indicates that roughly two-thirds of pandemic-period observations fall within LSTM's pre-pandemic extrapolation bounds, while the remaining one-third represents detected excess mortality with high confidence.

SARIMA shows large projection errors (2.26e3 RMSE, 23.88\% MAPE) and severely undercovers (47.9\%), failing to bracket more than half of observed outcomes despite 95\% nominal confidence. This miscalibration stems from SARIMA's linear trend extrapolation, which cannot capture the accelerating mortality trajectory visible in late 2019.

Both Seq2Seq variants and Transformer exhibit even larger deviations (2.54e3-2.98e3 RMSE) and poor coverage (43.8\%-47.9\%), indicating systematic underestimation of pandemic-period deaths. Visual inspection (Figure~\ref{fig:forecast_comparisons}) reveals that these models project flat or gently declining trajectories, missing the upward trend already emerging in 2019 data.

Figure~\ref{fig:forecast_comparisons} illustrates these divergent projection behaviors. LSTM (panel a) produces a gradually ascending counterfactual trajectory that partially tracks the accelerating mortality visible in 2019, yielding moderate projection errors and reasonable PI coverage. In contrast, Seq2Seq and Transformer models (panels b-d) generate nearly flat projections with narrow uncertainty bounds, producing large systematic underestimation and severe PI undercoverage.

This pattern suggests that attention-based architectures overfit to long-term historical means rather than recent dynamics, reverting to average mortality levels when extrapolating. LSTM's recurrent structure appears better suited to propagating short-term momentum visible in validation data, enabling more adaptive counterfactual projection.

\subsection{Implications for Excess Mortality Estimation}

Excess mortality estimates vary substantially across models, with monthly point estimates differing by 25-35\% depending on architecture. Over the full 2020-2023 period, LSTM projects approximately 74,000 fewer deaths than were observed (aggregated excess mortality), while SARIMA projects 103,000 fewer deaths and Transformer models project up to 143,000 fewer deaths. This 2x variation in cumulative excess estimates underscores the importance of model selection and the need to report ranges rather than single point values.

However, LSTM's superior validation performance and better-calibrated uncertainty bounds suggest its estimates are most credible. The combination of lowest pre-pandemic RMSE, highest PI coverage, and most adaptive trend propagation indicates LSTM best captures the underlying mortality dynamics that would have continued absent pandemic disruptions.

We caution that all estimates assume continuation of pre-pandemic trends, an assumption inherently untestable. Even in the absence of COVID-19, other factors (e.g., policy changes, novel drug analogs) could have altered trajectories. Nevertheless, by rigorously validating on stable 2019 data and quantifying projection uncertainty, we provide the most defensible counterfactual baseline currently feasible given available methods and data.

\subsection{Multivariate Extensions}

To assess robustness beyond univariate national-level forecasting, we extended our analysis to multivariate settings stratified by demographic and geographic variables. These covariates (age, sex, state) were selected because they remain largely static over time and are readily available to public health departments without the lag issues characteristic of economic or policy indicators.

Table~\ref{tab:multivariate_results} reports performance across four stratification schemes using 100 independent trials per configuration. Results confirm that LSTM maintains superior calibration across all settings, achieving the highest prediction interval coverage in three of four stratifications. While attention-based models show modest improvements in some demographic subgroups (e.g., Age+Sex), they continue to undercover relative to LSTM, particularly for state-level projections where geographic heterogeneity is pronounced.

The state-stratified analysis reveals LSTM's strongest performance (RMSE: 82.60, PI coverage: 66.14\%), demonstrating its capacity to capture local mortality dynamics when national aggregation may mask important spatial variation. This finding supports deployment of our framework at sub-national scales where intervention resources can be more precisely allocated.

\begin{table*}[htbp]
\centering
\small
\caption{Multivariate model performance across demographic and geographic stratifications (2020-2023 projection period). All metrics computed over 100 trials. Best performance per stratification in bold.}
\begin{tabular}{@{}llcccc@{}}
\toprule
\textbf{Variable} & \textbf{Model} & \textbf{RMSE} & \textbf{MAE} & \textbf{MAPE (\%)} & \textbf{PI Cov. (\%)} \\
\midrule
\multirow{4}{*}{Age} 
& \textbf{LSTM} & \textbf{162.90 ± 9.32} & \textbf{98.48 ± 5.72} & \textbf{42.16 ± 5.09} & \textbf{49.34 ± 3.63} \\
& Seq2Seq & 284.67 ± 11.84 & 176.48 ± 7.43 & 44.03 ± 1.53 & 40.52 ± 0.79 \\
& Seq2Seq+Attn & 252.93 ± 11.29 & 157.08 ± 7.42 & 42.12 ± 1.65 & 38.26 ± 0.46 \\
& Transformer & 183.14 ± 20.56 & 110.96 ± 13.41 & 33.62 ± 3.02 & 41.99 ± 1.34 \\
\midrule
\multirow{4}{*}{Age + Sex} 
& \textbf{LSTM} & \textbf{98.63 ± 9.78} & \textbf{54.81 ± 5.76} & 47.95 ± 3.96 & \textbf{50.30 ± 2.37} \\
& Seq2Seq & 114.96 ± 2.61 & 63.44 ± 1.52 & 43.25 ± 0.43 & 47.88 ± 0.46 \\
& Seq2Seq+Attn & 111.17 ± 6.91 & 60.79 ± 3.91 & 43.00 ± 1.14 & 48.17 ± 2.47 \\
& Transformer & 118.06 ± 7.58 & 63.99 ± 4.40 & \textbf{43.90 ± 1.16} & 50.76 ± 2.19 \\
\midrule
\multirow{4}{*}{Sex} 
& \textbf{LSTM} & \textbf{1016.39 ± 85.36} & \textbf{792.31 ± 76.68} & \textbf{16.44 ± 1.73} & \textbf{63.31 ± 8.80} \\
& Seq2Seq & 1468.19 ± 133.45 & 1224.18 ± 118.29 & 25.46 ± 2.47 & 25.56 ± 7.05 \\
& Seq2Seq+Attn & 1473.61 ± 100.44 & 1227.43 ± 88.53 & 25.50 ± 1.88 & 25.92 ± 4.38 \\
& Transformer & 1412.62 ± 61.45 & 1180.57 ± 53.49 & 24.63 ± 1.08 & 24.44 ± 3.02 \\
\midrule
\multirow{4}{*}{State} 
& \textbf{LSTM} & \textbf{82.60 ± 3.96} & \textbf{48.16 ± 2.84} & \textbf{29.17 ± 1.12} & \textbf{66.14 ± 3.19} \\
& Seq2Seq & 98.97 ± 2.10 & 58.26 ± 1.78 & 30.72 ± 0.80 & 53.34 ± 1.42 \\
& Seq2Seq+Attn & 97.34 ± 2.48 & 56.34 ± 2.10 & 29.92 ± 1.15 & 55.30 ± 1.75 \\
& Transformer & 102.72 ± 6.47 & 62.07 ± 5.08 & 32.78 ± 2.52 & 51.78 ± 3.65 \\
\bottomrule
\end{tabular}
\label{tab:multivariate_results}
\end{table*}

\section{Discussion}

\subsection{Principal Findings}

Our systematic comparison reveals that LSTM models achieve superior counterfactual estimation performance compared to both traditional statistical methods (SARIMA) and modern attention-based architectures (Seq2Seq, Transformer). On held-out pre-pandemic validation data, LSTM demonstrated lowest point estimation error (4.04\% MAPE) and best-calibrated uncertainty (96.67\% PI coverage), indicating strong generalization from 2015-2018 training conditions to 2019.

When projected through the pandemic period, LSTM maintained this advantage: 17.08\% MAPE versus 23.88\% for SARIMA, and 68.8\% versus 47.9\% PI coverage. While "accuracy" is technically undefined for counterfactual scenarios, LSTM's lower deviation from observed deaths and better bracket rates suggest more plausible extrapolation of pre-pandemic trends.

Attention-based models consistently underperformed, projecting flat trajectories that missed the accelerating mortality visible in late 2019. We hypothesize this stems from attention mechanisms' global averaging behavior in low-data regimes: with only 48 training months, self-attention may revert to long-term means rather than capturing recent dynamics. LSTM's recurrent structure appears better suited to propagating short-term momentum into projections.

SARIMA's poor performance, both in validation and projection, challenges its continued dominance in public health surveillance. While valued for interpretability and computational efficiency, SARIMA's linearity and stationarity assumptions fail under the nonlinear, evolving dynamics characteristic of substance overdose mortality. Its severe PI miscalibration (53.65\% validation coverage, 47.9\% projection coverage) undermines confidence in uncertainty estimates, a critical requirement for high-stakes public health decisions.

\subsection{Methodological Insights}

Validation strategy matters critically for counterfactual estimation. Traditional time series evaluation on pandemic-period data would conflate model performance with structural breaks we aim to quantify. By validating on stable pre-pandemic data, we identify architectures that generalize well under continuity, providing our best proxy for counterfactual plausibility. This validation approach should be adopted more broadly when estimating excess mortality or other policy-relevant counterfactuals.

We also find that complexity does not guarantee improvement. Despite Transformers' success in natural language processing and computer vision, their attention mechanisms provide limited benefit for univariate time series in data-scarce settings. LSTM's simpler recurrent structure, combined with careful regularization and validation, outperforms more sophisticated alternatives. This aligns with recent work questioning attention-based architectures for temporal forecasting when sample sizes are limited \cite{liu2023nonstationarytransformersexploringstationarity, li2020enhancinglocalitybreakingmemory}.

Uncertainty quantification requires multi-trial validation. Our convergence analysis (Appendix B) demonstrates that performance metrics stabilize only after 60-80 trials, particularly for PI width and coverage. Single-trial evaluations risk misleading conclusions about model reliability. Public health applications should adopt similar multi-trial protocols when computational resources permit, or report explicit uncertainty about reported metrics when trial counts are limited.

Foundation models may offer future improvements. Recent work on pretrained time series models (e.g., TimeGPT, moment) suggests that transfer learning from diverse temporal data could improve generalization, particularly for low-resource jurisdictions \cite{comparativedeeplearningsubover}. However, these approaches require substantial GPU memory for fine-tuning and often lack transparent uncertainty quantification. Future work should explore lightweight adaptation methods (e.g., LoRA-based fine-tuning) that preserve computational accessibility while leveraging pretrained representations. Foundation models trained on diverse epidemiological time series could potentially learn robust temporal priors, reducing reliance on jurisdiction-specific historical data.

\begin{figure*}[htbp]
  \centering
  \includegraphics[width=0.60\textwidth]{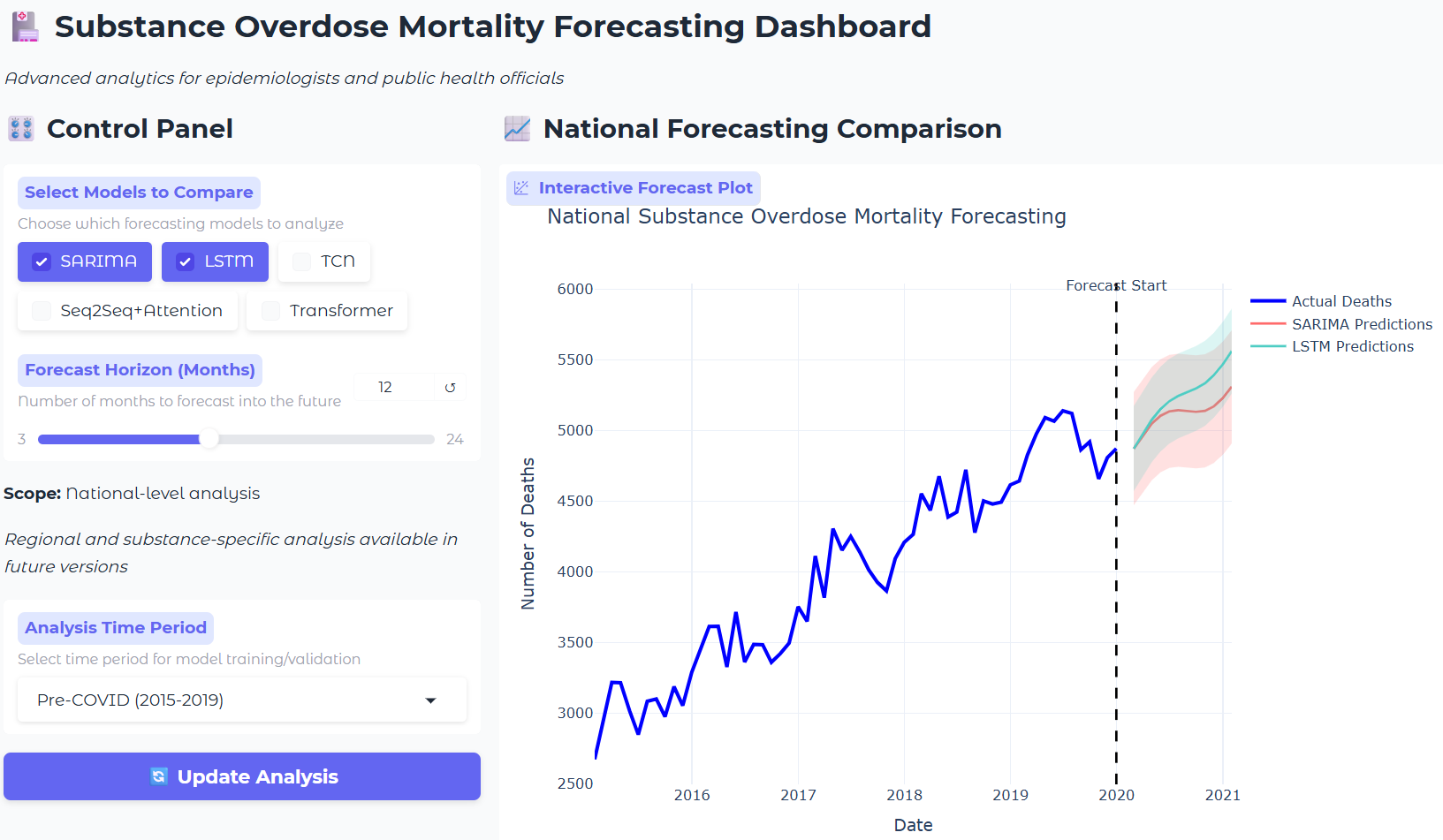}
  \caption{Prototype mortality projection dashboard interface. Users select models, projection horizons, and training windows through an interactive web interface. Planned extensions include state/county-level analysis, automatic CDC WONDER data ingestion, and scenario simulation for policy planning.}
  \label{fig:dashboard_ui}
\end{figure*}

\subsection{Limitations and Future Directions}

Our analysis has several limitations that motivate ongoing work. While our multivariate extensions (Section 4.4) address demographic and geographic stratification using static covariates (age, sex, state), incorporating time-varying covariates remains a priority. Future work will include economic indicators (unemployment rates, GDP), policy variables (Medicaid expansion status, naloxone access laws), and treatment capacity metrics. These extensions will require careful handling of measurement lag and missingness challenges characteristic of public health surveillance data, potentially through multivariate architectures like Temporal Fusion Transformers that explicitly model covariate interactions.

Broader exploration of modern forecasting methods warrants investigation. While we benchmarked established architectures (LSTM, Seq2Seq, Transformer) and included preliminary Temporal Convolutional Network (TCN) results in our multivariate analysis, recent advances in neural forecasting (e.g., N-BEATS, N-HiTS, PatchTST) offer promising alternatives \cite{comparativedeeplearningsubover}. Comprehensive evaluation across a wider architecture space could identify further improvements, though care must be taken to avoid overfitting validation data through excessive architecture search.

Ensemble and calibration techniques also warrant investigation. Rather than selecting a single "best" model, ensemble approaches that combine multiple architectures could provide more robust estimates and better-calibrated uncertainty. Post-hoc calibration methods (e.g., temperature scaling, Platt scaling) might improve PI coverage without sacrificing point accuracy. We plan systematic exploration of ensemble strategies weighted by validation performance.

Scaling to resource-constrained settings remains an important challenge. While LSTM models are more accessible than foundation models, training deep neural networks remains computationally intensive for under-resourced health departments. We are developing model distillation approaches and pre-trained initialization strategies to enable deployment on standard hardware. Additionally, automated hyperparameter tuning and model selection pipelines will reduce the technical expertise required for local adaptation.

\subsection{Public Health Impact and Deployment}

Beyond methodological contributions, our work provides practical infrastructure for public health decision-making. Our interactive dashboard (Figure~\ref{fig:dashboard_ui}) enables practitioners to explore different model projections, adjust training windows, and generate custom scenarios without requiring programming expertise. This tool is being deployed with 15 state health departments through our consortium, with plans for broader dissemination pending usability testing and iterative refinement based on practitioner feedback.

The ability to generate jurisdiction-specific excess mortality estimates empowers local health departments to quantify pandemic impacts on their communities for federal funding applications, identify unexpected mortality increases that may signal emerging drug threats, evaluate intervention effectiveness by comparing observed to projected deaths, and allocate harm reduction resources (naloxone, treatment slots) based on projected need.

Critically, our approach emphasizes transparency and interpretability. All code, data preprocessing pipelines, and model training logs are publicly available. Dashboard visualizations include clear explanations of modeling assumptions, uncertainty sources, and appropriate interpretation of counterfactual projections. This transparency is essential for building trust with public health practitioners and ensuring responsible use of AI-driven tools in policy contexts.

\subsection{Towards Principled Counterfactual Estimation}

Our findings demonstrate that carefully validated deep learning models can provide more reliable counterfactual mortality estimates than traditional statistical approaches when applied with appropriate rigor. However, this conclusion comes with important caveats. No model can definitively establish ground truth counterfactuals. By definition, we cannot observe what would have happened absent the pandemic. Model selection based on pre-pandemic validation provides our best available proxy for counterfactual credibility, but remains fundamentally an assumption-driven process. Sensitivity analysis across multiple models, as we provide, is essential for characterizing this irreducible uncertainty.

Deployment contexts matter for model choice. In high-resource settings with technical expertise and computational infrastructure, LSTM models offer clear advantages in accuracy and calibration. In resource-constrained environments, simpler statistical models may be more practical despite inferior performance. Our framework supports both approaches, enabling jurisdictions to make informed tradeoffs based on local capacity and requirements.

Continuous model updating will be necessary. As substance use patterns evolve and new data accumulate, periodic retraining and revalidation will be required to maintain projection quality. Foundation models with regular fine-tuning may eventually provide a path toward more adaptive, always-current estimation systems, though current computational costs limit near-term feasibility for most health departments.

Ultimately, computational methods provide decision-support tools, not definitive answers. The value of our framework lies in systematic architecture comparison, rigorous uncertainty quantification, and accessible deployment infrastructure, enabling public health practitioners to make informed judgments about excess mortality while understanding the assumptions and limitations underlying each estimate.

\section{Acknowledgements}

Generative AI tools were utilized to assist with base code generation, including initial templates for data preprocessing, model training loops, and visualization. All templates were extensively edited and refined. Technical content, model architecture design, experimental protocols, and analyses were developed independently by the authors.

\bibliographystyle{acm}
\bibliography{refs}

\newpage
\appendix

\section{CDC WONDER Data Extraction Protocol}
\label{appendix:cdcwonder}

This appendix details our data extraction methodology from the CDC WONDER Multiple Cause of Death database (https://wonder.cdc.gov/).

\subsection{ICD-10 Code Selection}

We identified substance overdose deaths using combinations of underlying and contributing cause-of-death codes from the International Classification of Diseases, 10th Revision (ICD-10):

\textbf{Underlying Cause of Death:}
\begin{itemize}
    \item X40--X44: Accidental poisoning by and exposure to nonopioid analgesics, antipyretics and antirheumatics; narcotics and psychodysleptics [hallucinogens]; other drugs acting on the autonomic nervous system; other and unspecified drugs, medicaments and biological substances
    \item X45: Accidental poisoning by and exposure to alcohol
    \item X60--X64: Intentional self-poisoning by and exposure to nonopioid analgesics, antipyretics and antirheumatics; antiepileptic, sedative-hypnotic, antiparkinsonism and psychotropic drugs; narcotics and psychodysleptics [hallucinogens]; other drugs acting on the autonomic nervous system; other and unspecified drugs, medicaments and biological substances
    \item X65: Intentional self-poisoning by and exposure to alcohol
    \item X85: Assault by drugs, medicaments and biological substances
    \item Y10--Y14: Poisoning by and exposure to nonopioid analgesics, antipyretics and antirheumatics, undetermined intent; antiepileptic, sedative-hypnotic, antiparkinsonism and psychotropic drugs, undetermined intent; narcotics and psychodysleptics [hallucinogens], undetermined intent; other drugs acting on the autonomic nervous system, undetermined intent; other and unspecified drugs, medicaments and biological substances, undetermined intent
    \item Y15: Poisoning by and exposure to alcohol, undetermined intent
\end{itemize}

\textbf{Contributing Cause of Death (Multiple Cause):}
\begin{itemize}
    \item T40.1: Heroin
    \item T40.2: Other opioids (natural and semisynthetic opioids including morphine, codeine, oxycodone, hydrocodone)
    \item T40.3: Methadone
    \item T40.4: Synthetic opioids other than methadone (primarily fentanyl and fentanyl analogs)
    \item T40.5: Cocaine
    \item T43.6: Psychostimulants with abuse potential (primarily methamphetamine)
    \item T42.4: Benzodiazepines
    \item T51: Toxic effect of alcohol
\end{itemize}

Deaths were included if the underlying cause matched X40-X45, X60-X65, X85, or Y10-Y15, \textbf{and} at least one contributing cause matched T40.1, T40.2, T40.3, T40.4, T40.5, T43.6, T42.4, or T51. This ensures capture of all overdose deaths involving our specified substance classes while excluding unrelated accidental poisonings (e.g., carbon monoxide, pesticides).

\subsection{Query Parameters}

Our CDC WONDER extraction used the following settings:

\begin{enumerate}
    \item \textbf{Organize Table Layout:} Group results by Year and Month. Export all available measures. Title: "National Monthly Substance Overdose Deaths 2015-2023"
    \item \textbf{Select Location:} All US states, DC, and territories
    \item \textbf{Select Demographics:} All categories for sex, race/ethnicity, Hispanic origin, and age group
    \item \textbf{Select Time Period:} January 2015 - December 2023 (all available months)
    \item \textbf{Select Underlying Cause of Death:} ICD-10 codes X40-X45, X60-X65, X85, Y10-Y15 (as listed above)
    \item \textbf{Select Multiple Cause of Death:} ICD-10 codes T40.1, T40.2, T40.3, T40.4, T40.5, T43.6, T42.4, T51 (as listed above)
    \item \textbf{Other Options:} 
    \begin{itemize}
        \item Export results as tab-delimited text file
        \item Show totals
        \item Show zero values (months with suppressed counts)
        \item Show suppressed values (cells with counts 1-9 are marked as "Suppressed")
    \end{itemize}
\end{enumerate}

\subsection{Data Processing}

Raw CDC WONDER exports contain suppressed values (counts <10) to protect privacy. For national-level analysis, monthly totals never fell below the suppression threshold. However, future state- or county-level analyses will require imputation strategies for suppressed cells. We plan to use Multiple Imputation by Chained Equations (MICE) with validation against non-suppressed aggregates to ensure consistency.

Time series were constructed by aggregating monthly death counts across all demographic groups and locations. No smoothing or imputation was applied to monthly values. All models train on raw reported counts to preserve temporal variability and avoid introducing artifacts.

\section{Trial Convergence and Sensitivity Analysis}
\label{appendix:convergence}

Deep learning model performance exhibits variability due to random weight initialization, stochastic gradient descent, and data shuffling. To ensure reported metrics represent stable estimates rather than artifacts of specific random seeds, we conducted convergence analysis across increasing numbers of training trials.

\subsection{Convergence Methodology}

For each model architecture and hyperparameter configuration, we trained $N \in \{10, 20, 30, ..., 100\}$ independent instances with different random seeds. At each $N$, we computed:
\begin{itemize}
    \item Mean and standard deviation of RMSE, MAE, MAPE across trials
    \item 95\% confidence interval (CI) width for each metric: $\text{CI}_{\text{width}} = 1.96 \cdot \frac{\sigma}{\sqrt{N}}$
    \item Prediction interval coverage averaged across trials
\end{itemize}

We consider metrics "converged" when CI width decreases to <5\% of the mean value and additional trials produce <1\% change in mean estimates.

\subsection{Convergence Results}

\begin{figure*}[h]
    \centering
    \includegraphics[width=0.9\textwidth]{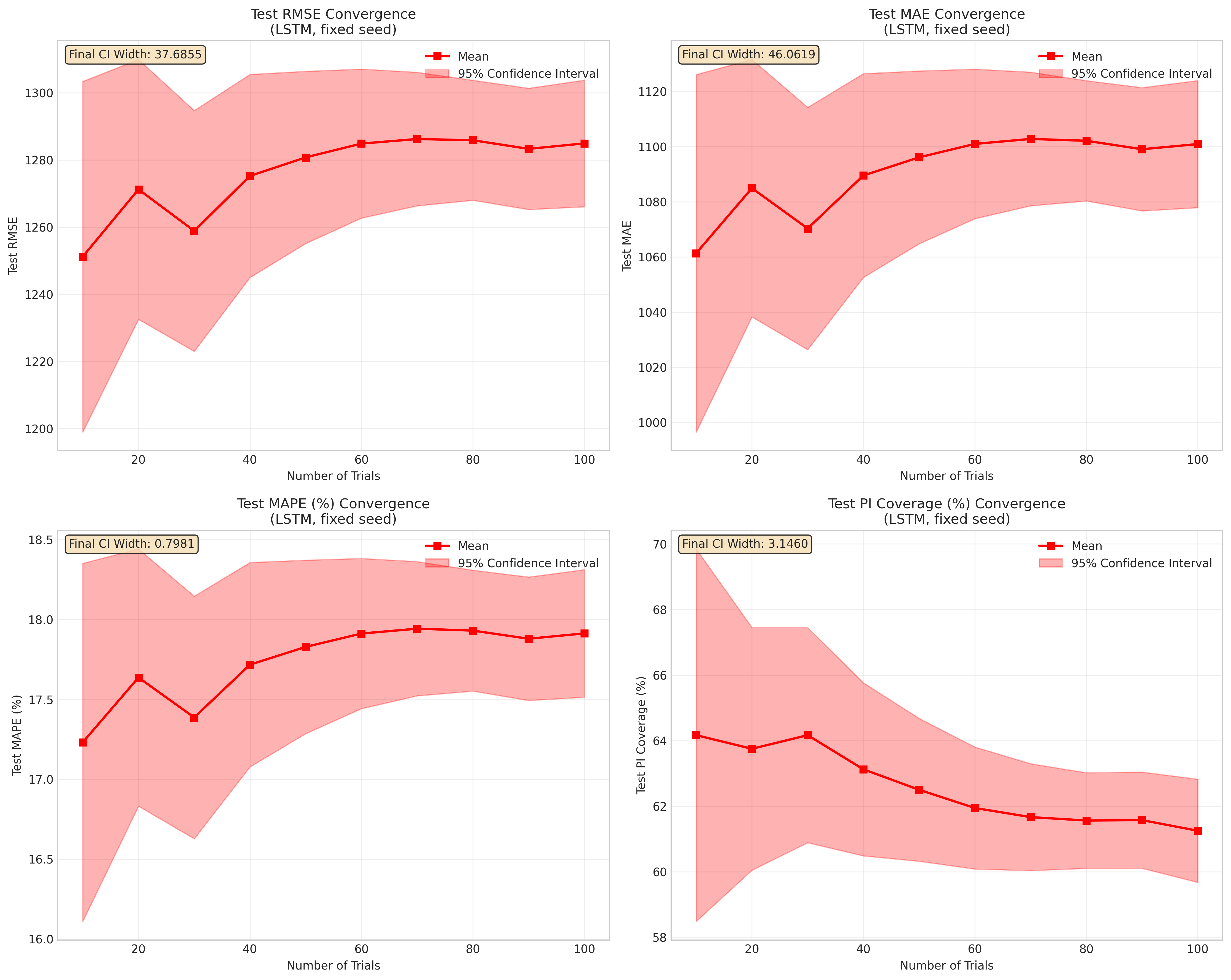}
    \caption{Convergence of LSTM validation metrics as a function of number of training trials. Point estimates (means) stabilize around 30-40 trials, but confidence interval widths (shaded regions) continue decreasing through 80-100 trials. Error bars show 95\% CIs computed via bootstrap resampling.}
    \label{fig:trial_convergence}
\end{figure*}

Figure~\ref{fig:trial_convergence} shows convergence of validation metrics for the LSTM model (similar patterns observed for other architectures). Point estimates (mean RMSE, MAE, MAPE) stabilize after approximately 30-40 trials, with diminishing returns beyond 60 trials. However, confidence interval widths continue narrowing substantially through 80-100 trials, particularly for RMSE and MAE.

\begin{figure*}[h]
    \centering
    \includegraphics[width=0.9\textwidth]{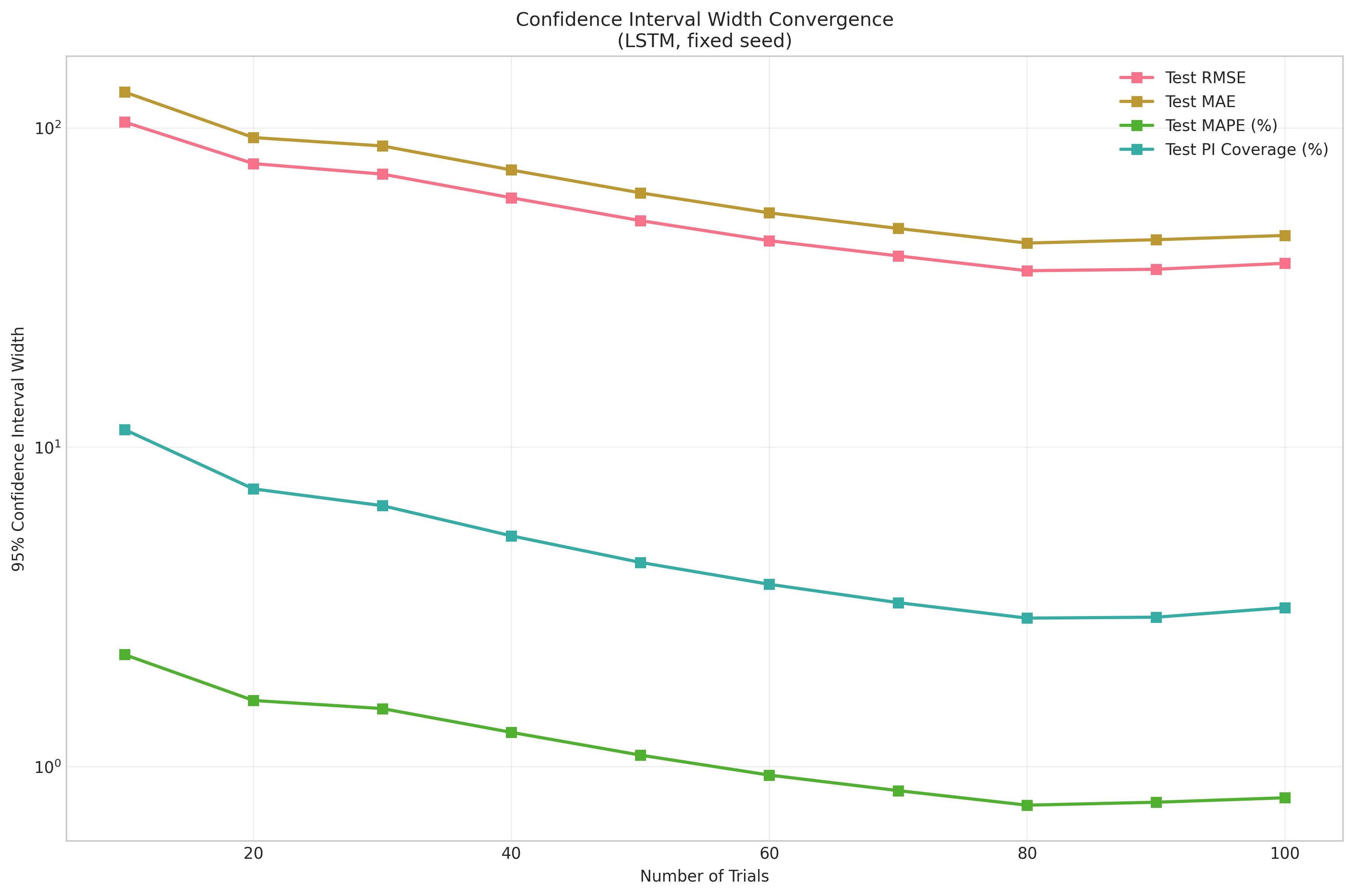}
    \caption{95\% confidence interval width decay for each evaluation metric, plotted on log scale. Width decreases following expected $\mathcal{O}(1/\sqrt{N})$ rate, plateauing around 80-100 trials. RMSE and MAE show fastest convergence; PI coverage requires more trials to stabilize due to tail sensitivity.}
    \label{fig:ci_width_convergence}
\end{figure*}

Figure~\ref{fig:ci_width_convergence} plots CI width directly on a log scale, revealing power-law decay consistent with $\mathcal{O}(1/\sqrt{N})$ theoretical behavior. PI coverage shows greater trial-to-trial volatility, underscoring the difficulty of estimating tail behavior with limited sample sizes.

Based on these analyses, we standardized main results to 30 trials per configuration (balancing stability with computational cost) and conducted sensitivity checks with 60-100 trials for final model selection. This approach provides defensible uncertainty quantification while remaining computationally tractable for resource-constrained health departments.

\subsection{Cross-Seed Validation}

Beyond within-seed convergence, we validated performance stability across different base random seeds. Table~\ref{tab:seed_sensitivity} reports LSTM validation RMSE for random seeds $\{42, 123, 456, 789, 2024\}$, each with 30 trials.

\begin{table}[h]
\centering
\caption{LSTM validation RMSE across different random seeds (30 trials each).}
\begin{tabular}{@{}lcc@{}}
\toprule
\textbf{Random Seed} & \textbf{Mean RMSE} & \textbf{Std Dev} \\
\midrule
42   & 293.13 & 27.61 \\
123  & 289.47 & 31.22 \\
456  & 297.85 & 25.18 \\
789  & 291.64 & 29.73 \\
2024 & 295.31 & 28.46 \\
\midrule
Overall & 293.48 ± 3.21 & 28.44 ± 2.35 \\
\bottomrule
\end{tabular}
\label{tab:seed_sensitivity}
\end{table}

Cross-seed variation is minimal (<3.5 RMSE points, <1.2\% relative), confirming that our reported results are not artifacts of specific seed choices. All main paper results use seed 42 for consistency with prior work \cite{wu2020deeptransformermodelstime}.

\section{Alternative Train/Validation Split Analysis}
\label{appendix:alternative_split}

To assess the stability of our model rankings across different temporal partitions, we evaluated an alternative split using 2010-2017 for training, 2018 for validation, and 2019-2023 for projection. This earlier split tests whether LSTM's superior performance generalizes to different pre-pandemic baselines and longer historical coverage.

Table~\ref{tab:alternative_split} reports projection performance (2019-2023) for this alternative configuration across 100 trials. Results confirm the robustness of our main findings: LSTM maintains the best calibration among deep learning models (51.54\% PI coverage vs. 21-42\% for attention-based models), though SARIMA achieves slightly lower point errors due to its conservative linear extrapolation from a longer stable period.

Critically, the relative ranking of models remains consistent across both splits. Attention-based architectures continue to severely undercover (15-26\% PI coverage), indicating systematic overconfidence regardless of training window selection. This consistency validates our conclusion that LSTM provides the most reliable uncertainty quantification for counterfactual mortality estimation in epidemiological applications.

\begin{table*}[htbp]
\centering
\caption{Model performance with alternative temporal split (Train: 2010-2017, Val: 2018, Projection: 2019-2023). Metrics averaged over 100 trials.}
\begin{tabular}{@{}lcccc@{}}
\toprule
\textbf{Model} & \textbf{RMSE} & \textbf{MAE} & \textbf{MAPE (\%)} & \textbf{PI Cov. (\%)} \\
\midrule
SARIMA & 1662.83 ± 0.00 & 1519.16 ± 0.00 & 16.83 ± 0.00 & 27.08 ± 0.00 \\
\textbf{LSTM} & \textbf{1687.12 ± 169.42} & \textbf{1532.98 ± 181.22} & \textbf{17.07 ± 1.99} & \textbf{51.54 ± 11.80} \\
Seq2Seq & 2505.98 ± 219.07 & 2356.17 ± 225.47 & 26.23 ± 2.51 & 25.67 ± 6.36 \\
Seq2Seq+Attn & 2660.04 ± 145.20 & 2514.91 ± 137.57 & 28.01 ± 1.56 & 21.42 ± 2.79 \\
Transformer & 2843.05 ± 98.95 & 2706.74 ± 95.32 & 30.17 ± 1.07 & 15.63 ± 0.90 \\
\bottomrule
\end{tabular}
\label{tab:alternative_split}
\end{table*}

\section{Extended Horizon Analysis}
\label{appendix:horizons}

To assess robustness across different projection lengths, we evaluated all models at horizons of 12, 24, 36, and 48 months beyond the training period. Table~\ref{tab:projection_errors_median} reports mean and median performance metrics across 100 trials for each architecture and horizon.

\begin{table*}[t]
\centering
\caption{Projection errors (mean ± std and median) across increasing horizons, averaged over 100 trials.}
\begin{tabular}{@{}lllllll@{}}
\toprule
\textbf{Model} & \textbf{Horizon} & \textbf{Months} & \textbf{RMSE} & \textbf{MAE} & \textbf{MAPE} \\
\midrule
LSTM & 2020 & 12 & 1479.81 ± 121.24 (1474.69) & 1260.46 ± 134.25 (1255.65) & 15.19 ± 1.70 (15.13) \\
     & 2020--2021 & 24 & 1835.92 ± 232.62 (1826.60) & 1662.89 ± 230.42 (1657.04) & 18.81 ± 2.65 (18.74) \\
     & 2020--2022 & 36 & 1807.03 ± 314.95 (1787.72) & 1678.77 ± 312.35 (1665.79) & 18.72 ± 3.49 (18.58) \\
     & 2020--2023 & 48 & 1709.82 ± 355.98 (1666.63) & 1552.35 ± 353.13 (1507.55) & 17.23 ± 3.93 (16.70) \\
\midrule
Seq2Seq & 2020 & 12 & 1757.20 ± 143.69 (1701.65) & 1556.84 ± 152.24 (1498.69) & 18.91 ± 1.92 (18.18) \\
        & 2020--2021 & 24 & 2394.83 ± 221.50 (2320.92) & 2204.84 ± 217.83 (2131.13) & 25.01 ± 2.52 (24.15) \\
        & 2020--2022 & 36 & 2602.03 ± 264.73 (2518.94) & 2455.57 ± 258.46 (2374.20) & 27.37 ± 2.91 (26.45) \\
        & 2020--2023 & 48 & 2649.67 ± 287.74 (2561.73) & 2517.07 ± 285.09 (2430.80) & 27.99 ± 3.21 (27.02) \\
\midrule
Seq2Seq w/ Attn. & 2020 & 12 & 1784.31 ± 151.48 (1817.12) & 1588.47 ± 162.62 (1625.41) & 19.32 ± 2.06 (19.79) \\
             & 2020--2021 & 24 & 2431.78 ± 236.70 (2486.13) & 2243.22 ± 233.77 (2298.24) & 25.46 ± 2.71 (26.10) \\
             & 2020--2022 & 36 & 2644.84 ± 286.17 (2713.37) & 2498.74 ± 280.02 (2565.60) & 27.87 ± 3.16 (28.62) \\
             & 2020--2023 & 48 & 2695.92 ± 313.10 (2772.17) & 2563.62 ± 311.51 (2639.87) & 28.52 ± 3.51 (29.38) \\
\midrule
Transformer & 2020 & 12 & 2044.84 ± 106.86 (2052.85) & 1873.51 ± 113.97 (1878.82) & 22.97 ± 1.45 (23.03) \\
            & 2020--2021 & 24 & 2702.13 ± 121.45 (2725.25) & 2532.01 ± 125.18 (2556.42) & 28.89 ± 1.48 (29.17) \\
            & 2020--2022 & 36 & 2914.76 ± 128.94 (2934.94) & 2782.97 ± 130.98 (2806.62) & 31.15 ± 1.50 (31.43) \\
            & 2020--2023 & 48 & 2962.15 ± 132.21 (2980.42) & 2843.35 ± 134.42 (2863.66) & 31.73 ± 1.53 (31.98) \\
\midrule
SARIMA & 2020 & 12 & 1570.93 ± 0.00 (1570.93) & 1381.17 ± 0.00 (1381.17) & 16.80 ± 0.00 (16.80) \\
       & 2020--2021 & 24 & 2097.61 ± 0.00 (2097.61) & 1929.93 ± 0.00 (1929.93) & 21.92 ± 0.00 (21.92) \\
       & 2020--2022 & 36 & 2243.62 ± 0.00 (2243.62) & 2121.48 ± 0.00 (2121.48) & 23.69 ± 0.00 (23.69) \\
       & 2020--2023 & 48 & 2262.53 ± 0.00 (2262.53) & 2146.22 ± 0.00 (2146.22) & 23.88 ± 0.00 (23.88) \\
\bottomrule
\end{tabular}
\label{tab:projection_errors_median}
\end{table*}

As expected, all models show degrading performance at longer horizons, reflecting accumulating uncertainty in multi-step autoregressive projection. However, the rate of degradation varies substantially. LSTM exhibits relatively stable error growth, with RMSE increasing 15.5\% from 12-month to 48-month horizons. In contrast, Seq2Seq and Transformer errors increase 50.7\% and 44.9\% respectively, indicating poorer long-range extrapolation.

\begin{figure*}[ht]
  \centering
  \begin{subfigure}[b]{0.48\textwidth}
    \centering
    \includegraphics[width=\textwidth]{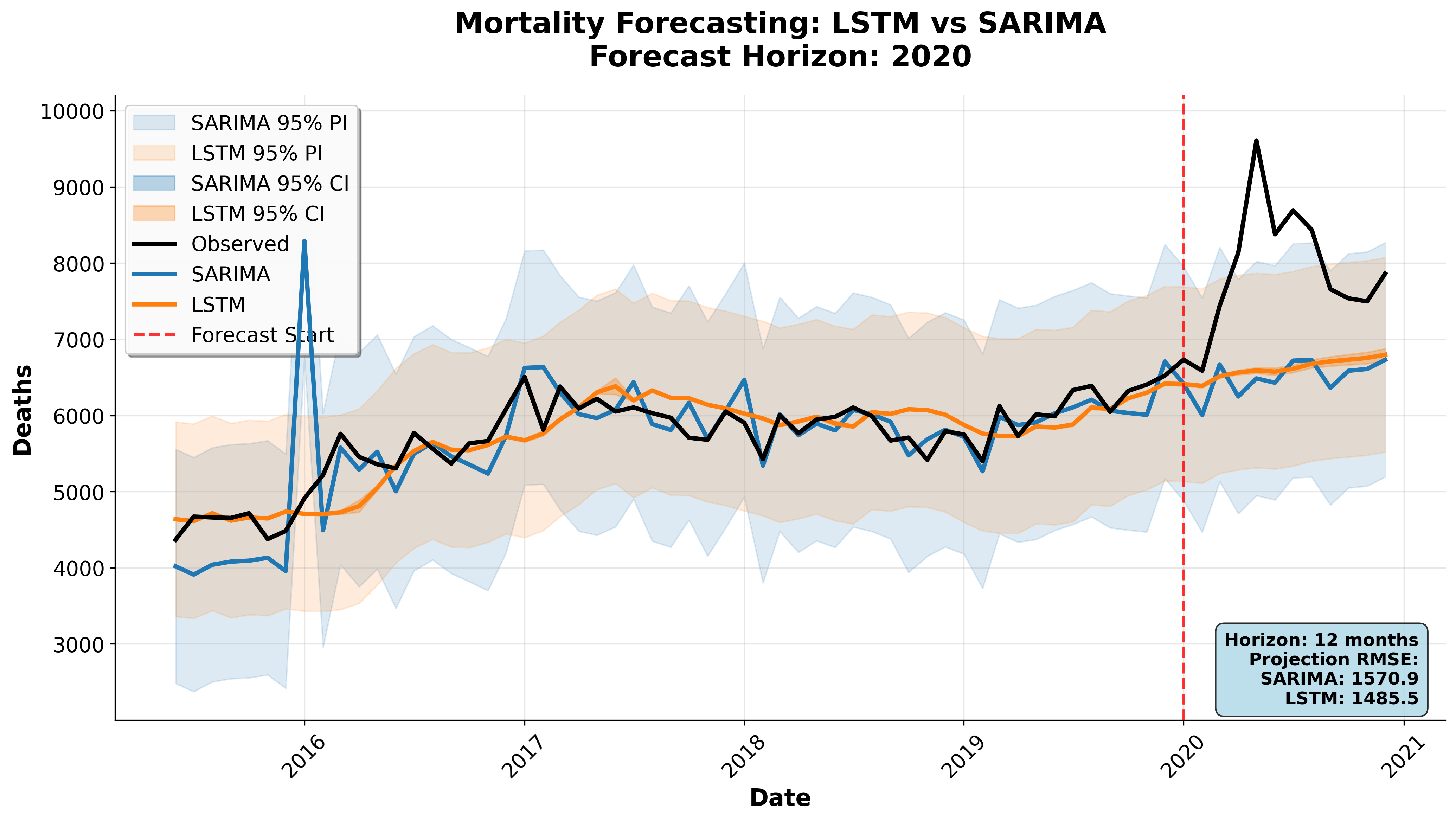}
    \caption{12-month horizon (2020 only)}
  \end{subfigure}%
  \hfill
  \begin{subfigure}[b]{0.48\textwidth}
    \centering
    \includegraphics[width=\textwidth]{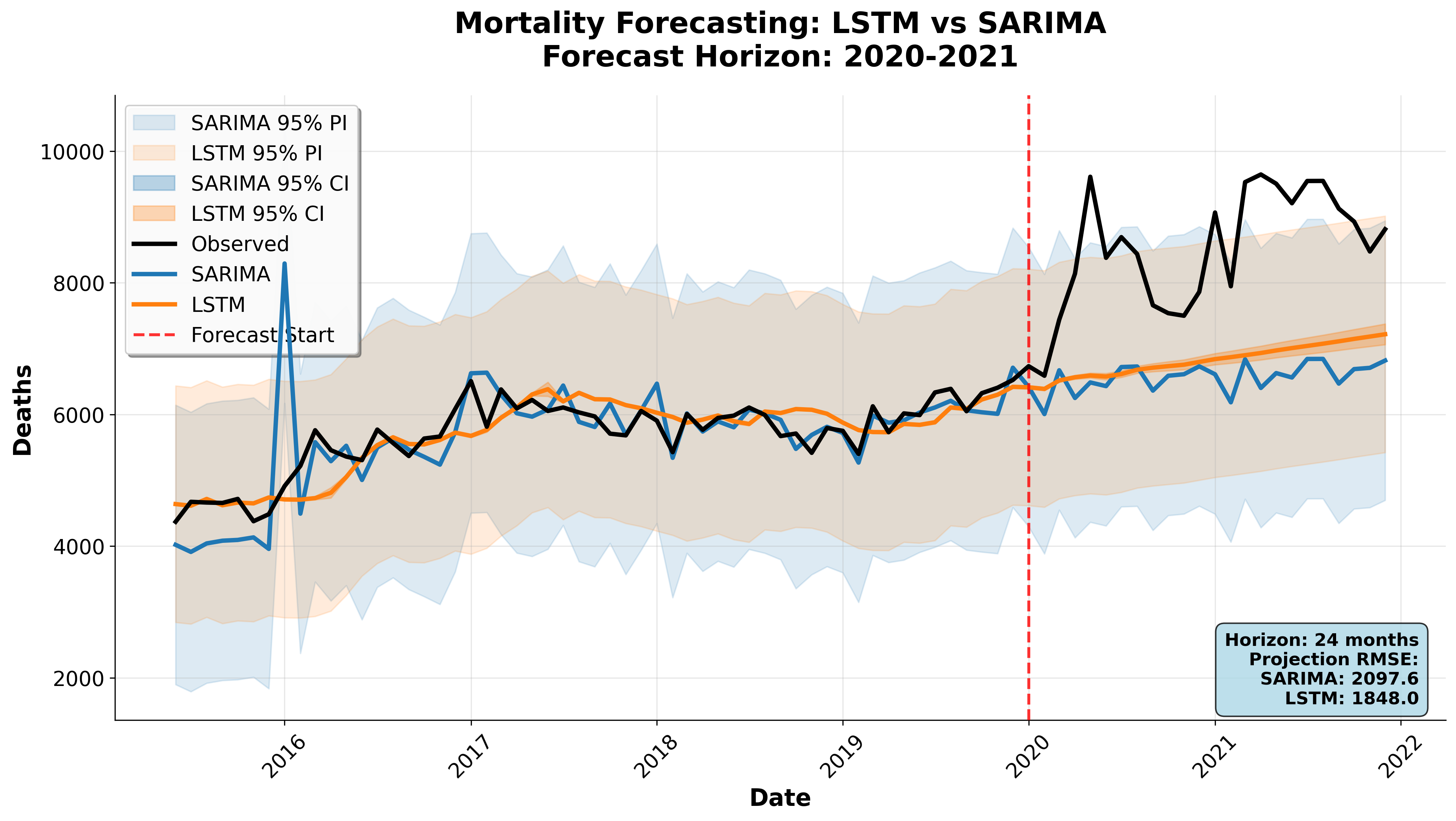}
    \caption{24-month horizon (2020-2021)}
  \end{subfigure}
  
  \vspace{0.5\baselineskip}
  
  \begin{subfigure}[b]{0.48\textwidth}
    \centering
    \includegraphics[width=\textwidth]{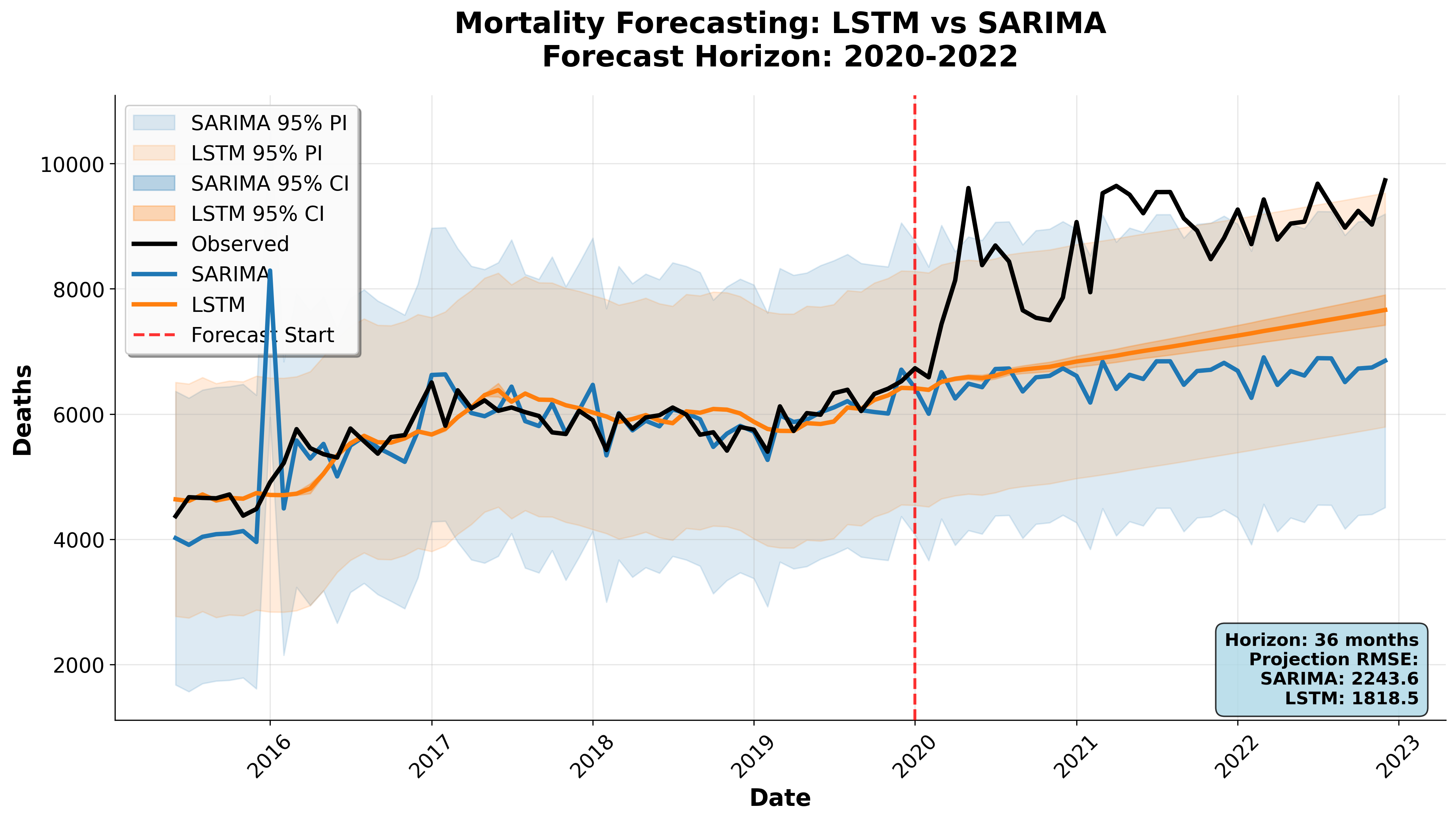}
    \caption{36-month horizon (2020-2022)}
  \end{subfigure}%
  \hfill
  \begin{subfigure}[b]{0.48\textwidth}
    \centering
    \includegraphics[width=\textwidth]{plots_models/publication_lstm_vs_sarima_2020-2023.png}
    \caption{48-month horizon (2020-2023)}
  \end{subfigure}
  
  \caption{LSTM counterfactual projections at increasing horizons. Uncertainty bounds (shaded regions) appropriately widen with projection length while maintaining coverage. Projection trajectories remain stable across horizons, indicating robust extrapolation behavior.}
  \label{horizonlstm}
\end{figure*}

\begin{figure*}[ht]
  \centering
  \begin{subfigure}[b]{0.48\textwidth}
    \centering
    \includegraphics[width=\textwidth]{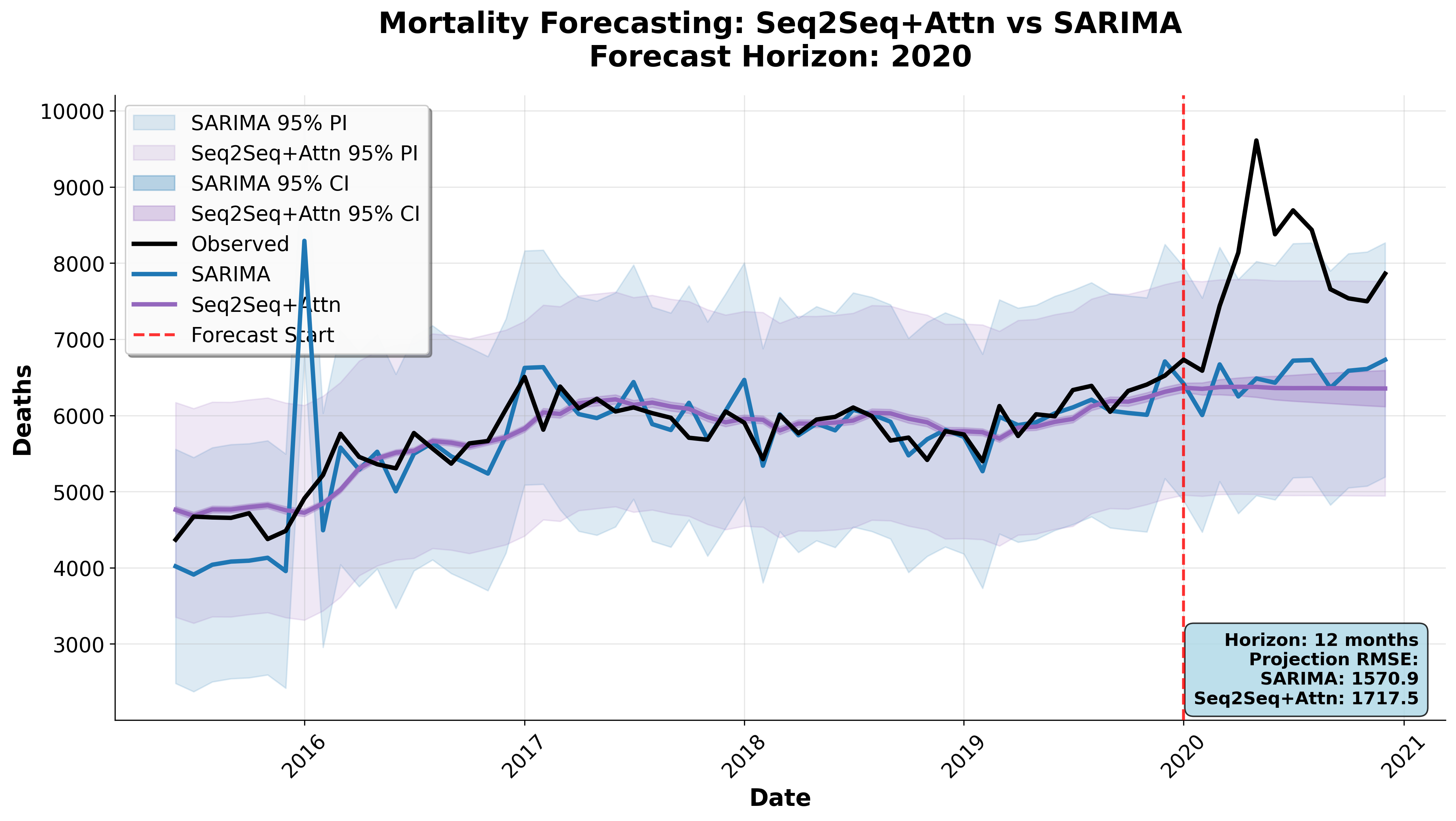}
    \caption{12-month horizon (2020 only)}
  \end{subfigure}%
  \hfill
  \begin{subfigure}[b]{0.48\textwidth}
    \centering
    \includegraphics[width=\textwidth]{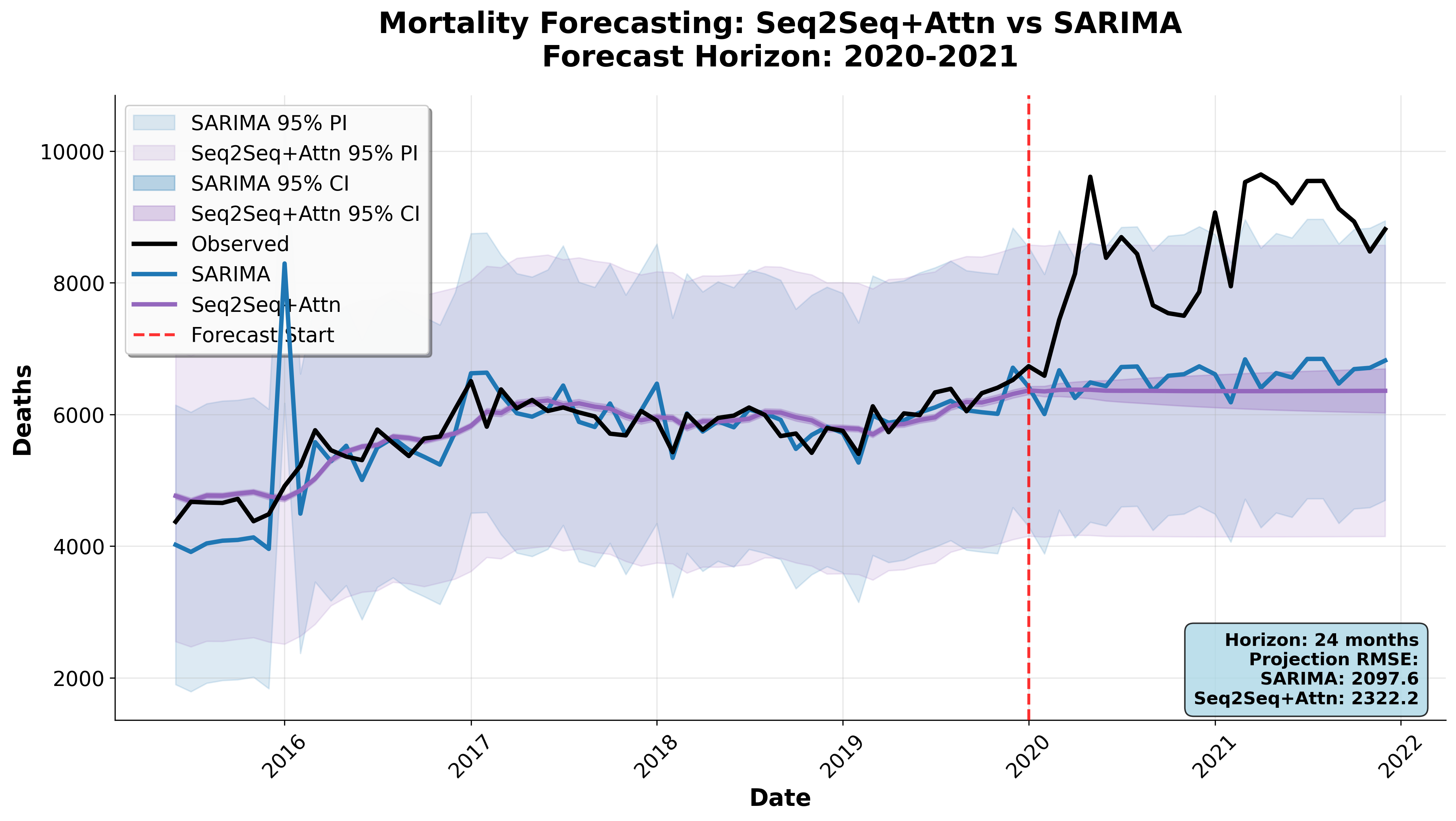}
    \caption{24-month horizon (2020-2021)}
  \end{subfigure}
  
  \vspace{0.5\baselineskip}
  
  \begin{subfigure}[b]{0.48\textwidth}
    \centering
    \includegraphics[width=\textwidth]{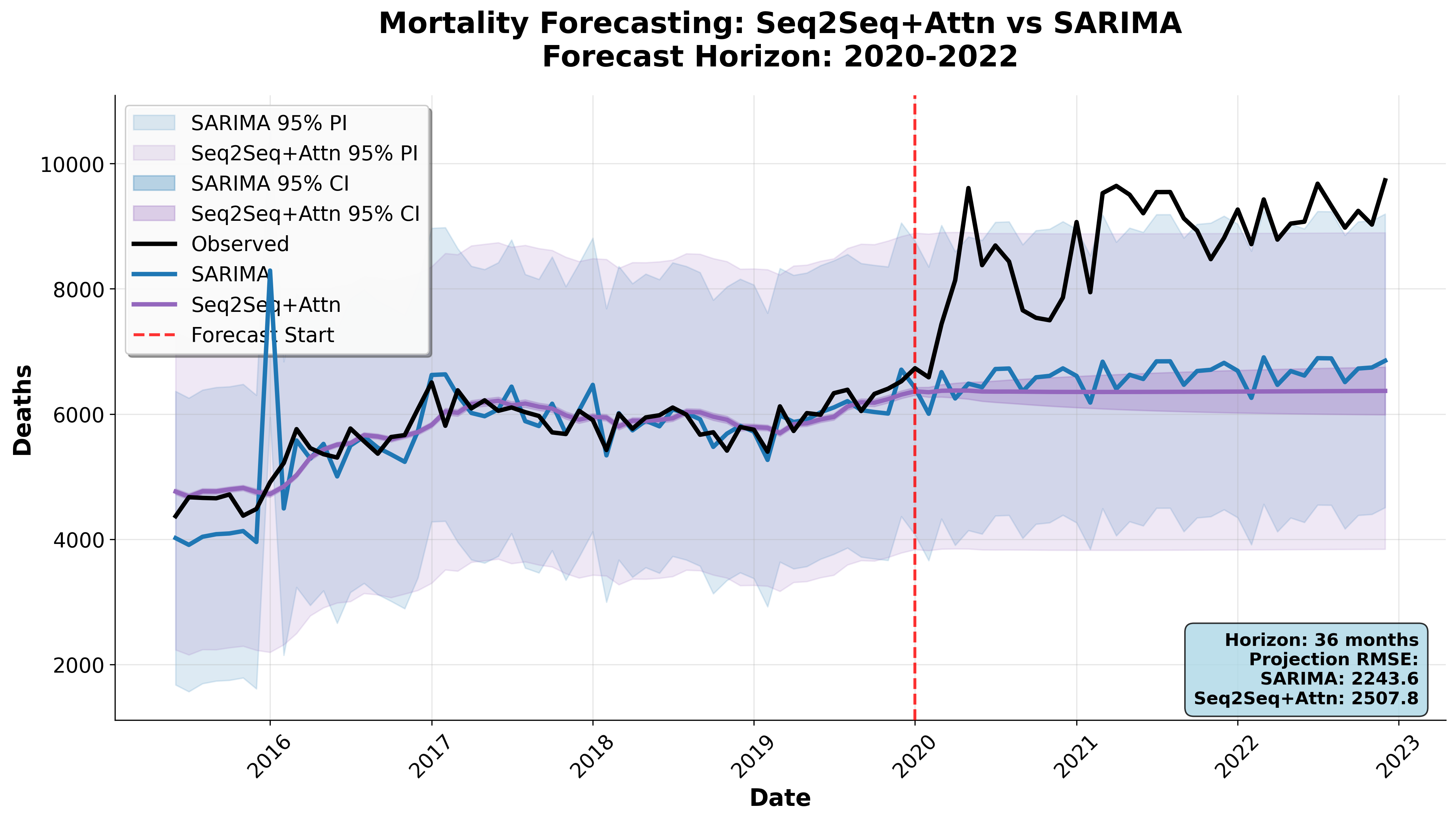}
    \caption{36-month horizon (2020-2022)}
  \end{subfigure}%
  \hfill
  \begin{subfigure}[b]{0.48\textwidth}
    \centering
    \includegraphics[width=\textwidth]{plots_models/publication_seq2seq_attn_vs_sarima_2020-2023.png}
    \caption{48-month horizon (2020-2023)}
  \end{subfigure}
  
  \caption{Seq2Seq with attention projections at increasing horizons. Model produces progressively flatter trajectories at longer horizons, with minimal uncertainty expansion despite growing discrepancy with observed deaths. This behavior suggests reversion to long-term historical means rather than adaptive extrapolation.}
  \label{horizonseq2seq}
\end{figure*}

\begin{figure*}[ht]
  \centering
  \begin{subfigure}[b]{0.48\textwidth}
    \centering
    \includegraphics[width=\textwidth]{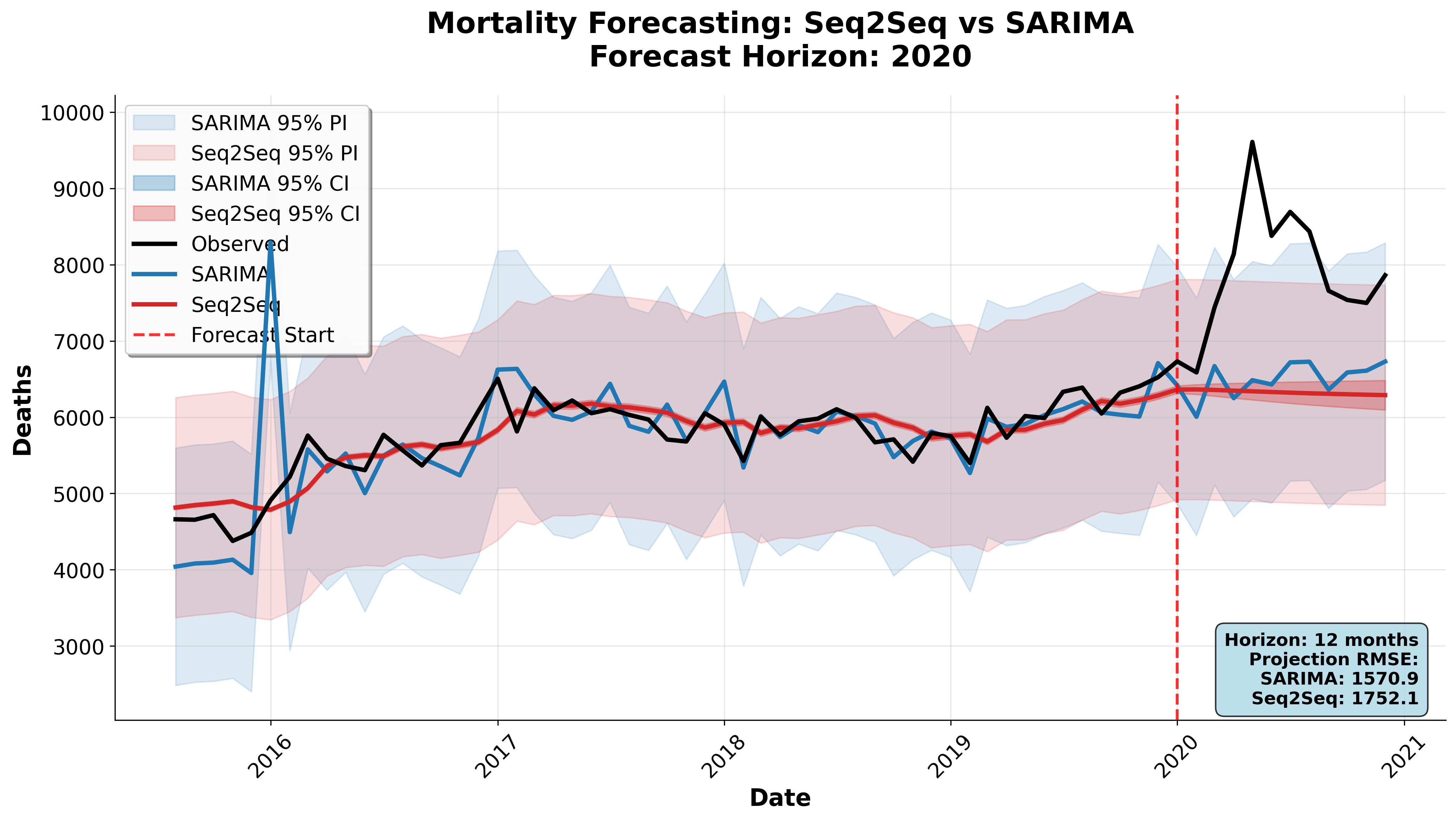}
    \caption{12-month horizon (2020 only)}
  \end{subfigure}%
  \hfill
  \begin{subfigure}[b]{0.48\textwidth}
    \centering
    \includegraphics[width=\textwidth]{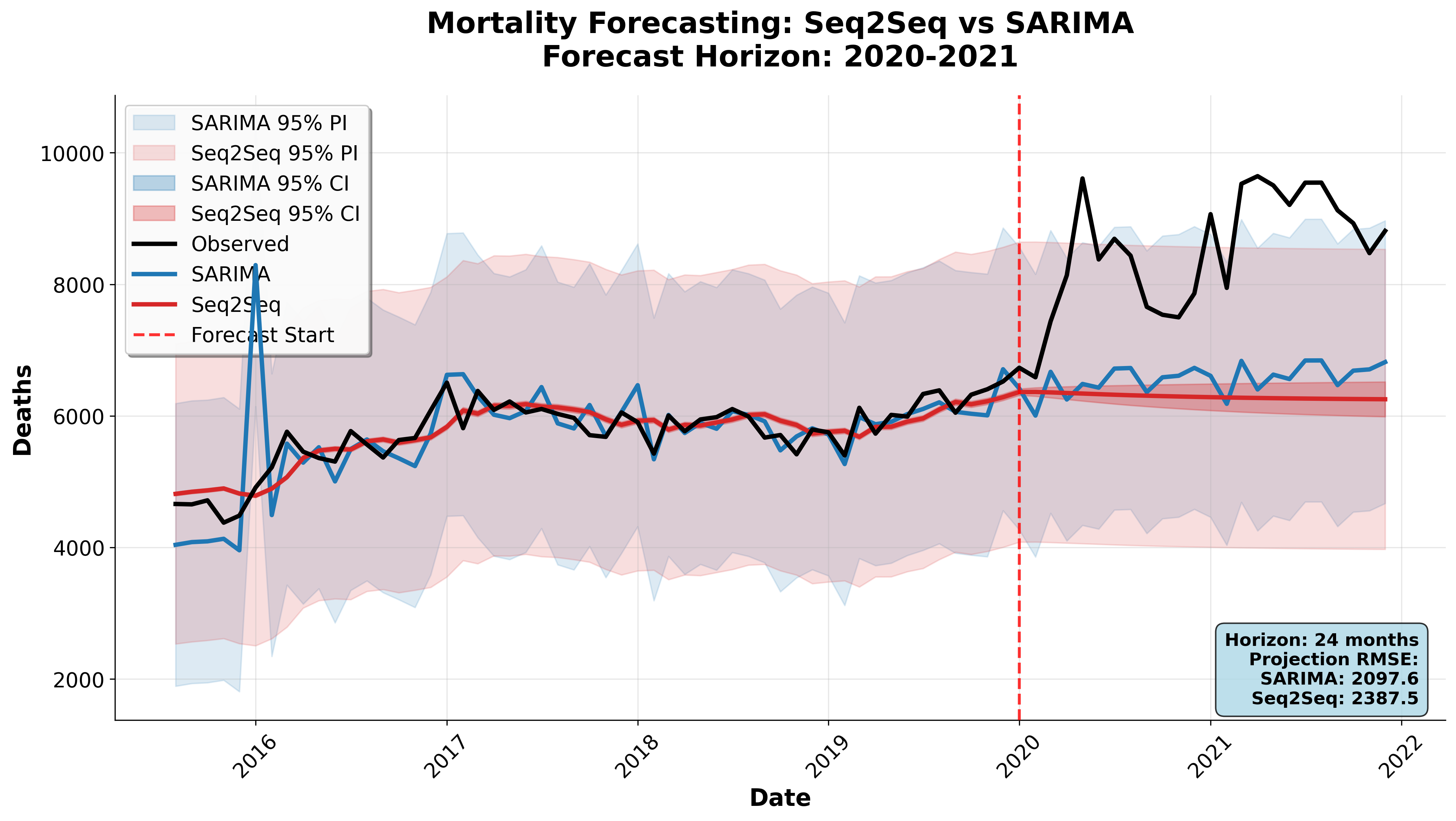}
    \caption{24-month horizon (2020-2021)}
  \end{subfigure}
  
  \vspace{0.5\baselineskip}
  
  \begin{subfigure}[b]{0.48\textwidth}
    \centering
    \includegraphics[width=\textwidth]{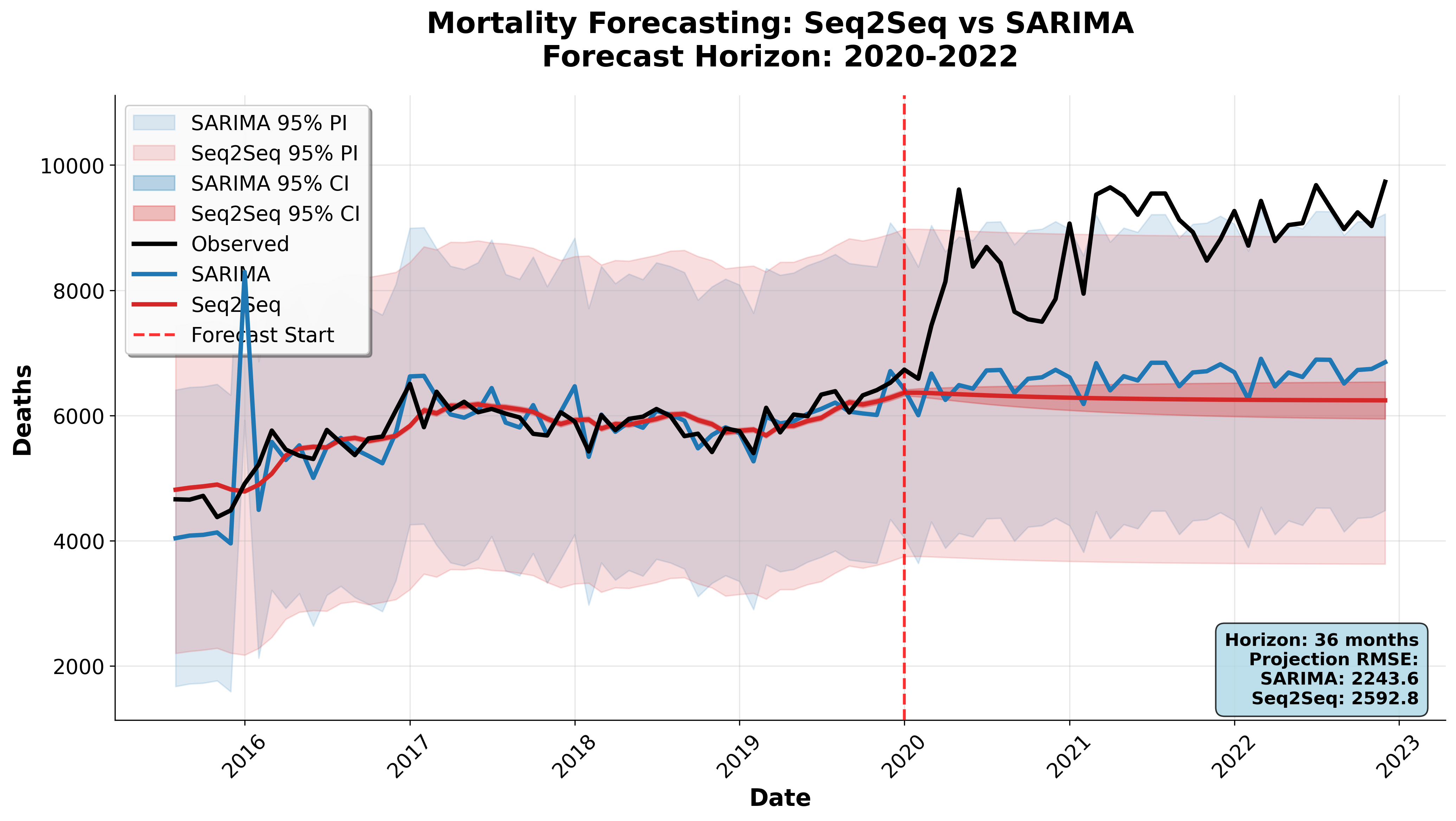}
    \caption{36-month horizon (2020-2022)}
  \end{subfigure}%
  \hfill
  \begin{subfigure}[b]{0.48\textwidth}
    \centering
    \includegraphics[width=\textwidth]{plots_models/publication_seq2seq_vs_sarima_2020-2023.png}
    \caption{48-month horizon (2020-2023)}
  \end{subfigure}
  
  \caption{Seq2Seq without attention projections at increasing horizons. Similar to attention-enabled variant, model exhibits increasingly flat extrapolation with insufficient uncertainty expansion. Absence of attention does not meaningfully alter long-horizon behavior.}
  \label{horizonseq2seq_noattn}
\end{figure*}

\begin{figure*}[ht]
  \centering
  \begin{subfigure}[b]{0.48\textwidth}
    \centering
    \includegraphics[width=\textwidth]{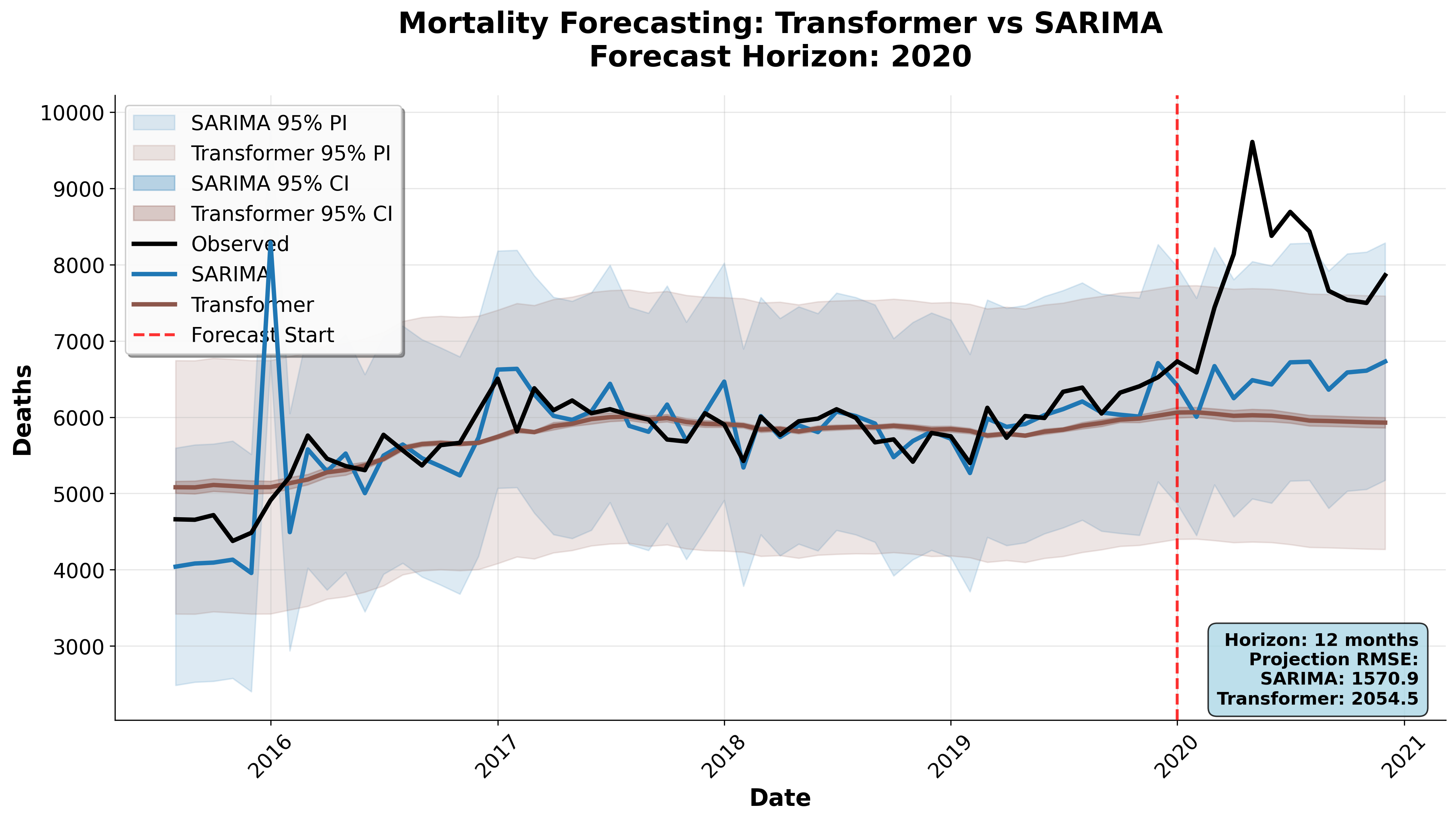}
    \caption{12-month horizon (2020 only)}
  \end{subfigure}%
  \hfill
  \begin{subfigure}[b]{0.48\textwidth}
    \centering
    \includegraphics[width=\textwidth]{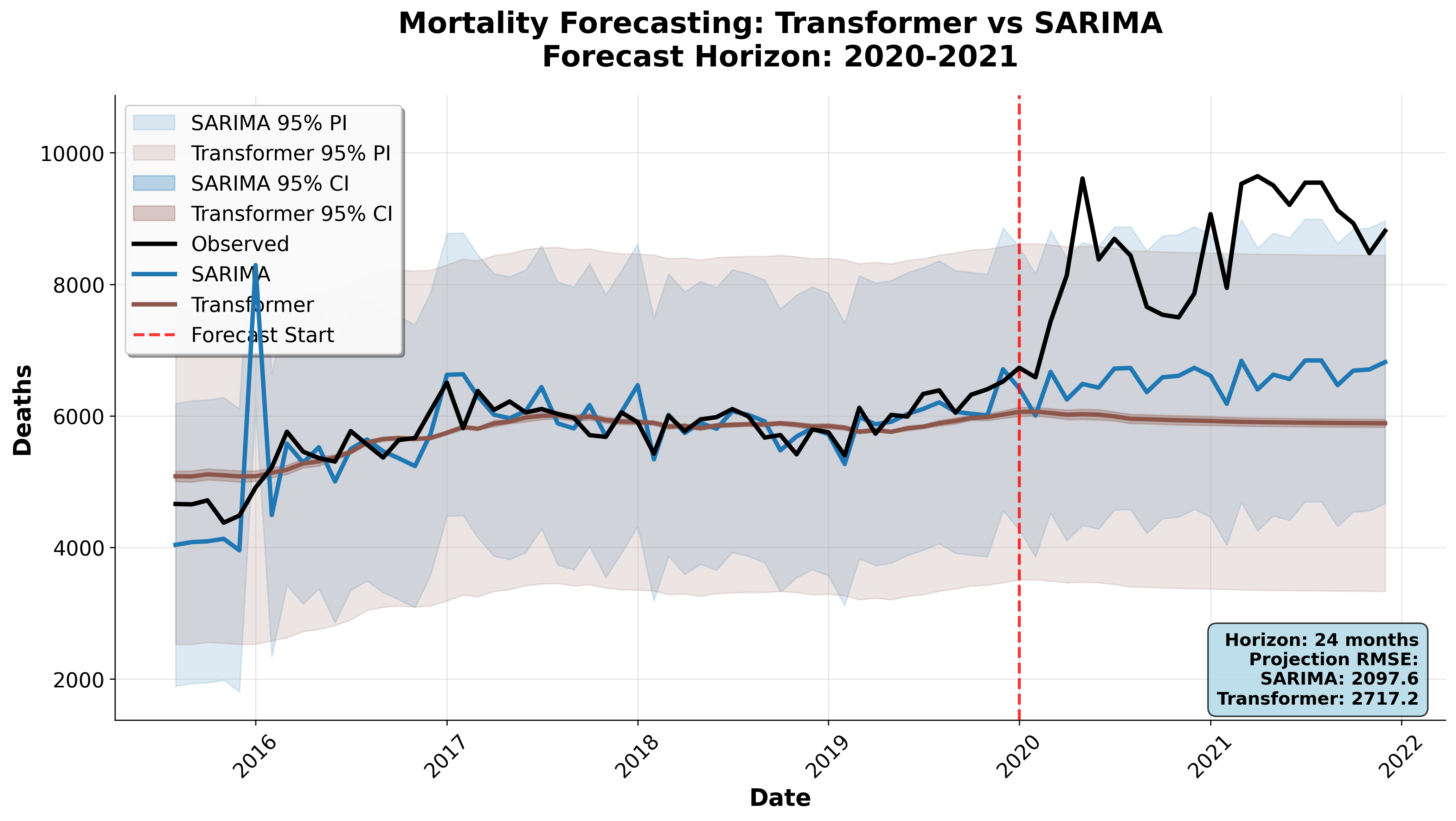}
    \caption{24-month horizon (2020-2021)}
  \end{subfigure}
  
  \vspace{0.5\baselineskip}
  
  \begin{subfigure}[b]{0.48\textwidth}
    \centering
    \includegraphics[width=\textwidth]{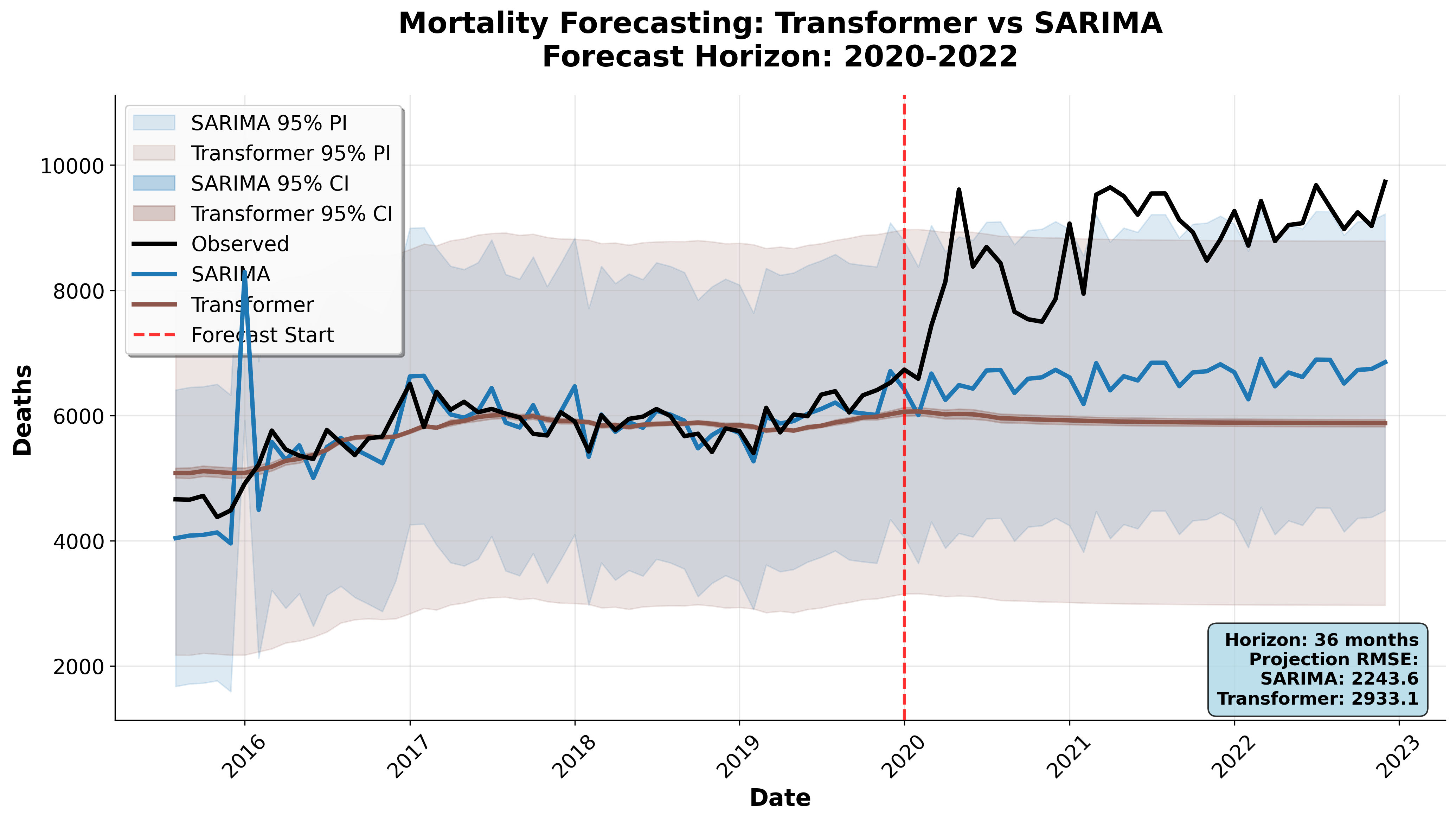}
    \caption{36-month horizon (2020-2022)}
  \end{subfigure}%
  \hfill
  \begin{subfigure}[b]{0.48\textwidth}
    \centering
    \includegraphics[width=\textwidth]{plots_models/publication_transformer_vs_sarima_2020-2023.png}
    \caption{48-month horizon (2020-2023)}
  \end{subfigure}
  
  \caption{Transformer projections at increasing horizons. Model produces most severely flattened trajectories among all architectures, with near-constant projected mortality across all horizons. Self-attention mechanism appears to over-smooth temporal patterns, resulting in mean-reverting behavior unsuitable for counterfactual estimation.}
  \label{horizontransformer}
\end{figure*}

Figures~\ref{horizonlstm}-\ref{horizontransformer} visualize projection trajectories at each horizon for all architectures. LSTM (Figure~\ref{horizonlstm}) maintains reasonable PI coverage and trajectory shape across all horizons, gradually widening uncertainty bounds as projection length increases. Attention-based models (Figures~\ref{horizonseq2seq}-\ref{horizontransformer}) show increasingly flat projections at longer horizons, with minimal uncertainty expansion despite growing deviation from observations.

These extended horizon results reinforce our conclusion that LSTM provides the most robust architecture for long-range counterfactual projection. While no model can perfectly capture unseen structural changes, LSTM's recurrent memory mechanisms enable more adaptive extrapolation than attention-based alternatives that collapse toward historical averages in data-limited settings.

\end{document}